\documentclass{article} 
\usepackage{iclr2019_conference,times}
\usepackage{appendix}

\usepackage{amsmath,amsfonts,bm}

\def\eqref#1{equation~\ref{#1}}

\def\1{\bm{1}}

\DeclareMathAlphabet{\mathsfit}{\encodingdefault}{\sfdefault}{m}{sl}
\SetMathAlphabet{\mathsfit}{bold}{\encodingdefault}{\sfdefault}{bx}{n}

\usepackage{wrapfig}
\usepackage{hyperref}
\usepackage{url}
\usepackage{microtype}
\usepackage{graphicx}
\usepackage{amssymb}
\usepackage{amsmath}
\usepackage{amsthm}
\usepackage[nomargin,inline,marginclue,draft]{fixme}
\usepackage{subfigure}
\usepackage{xfrac}
\usepackage{booktabs} 
\usepackage{tikz}

\newtheorem{remark}{Lemma}
\usepackage{thmtools}
\usepackage{thm-restate}

\title{Clique Pooling for Graph Classification}

\author{Enxhell Luzhnica\thanks{Equal contribution},\ \ Ben Day\footnotemark[1]\ \ \& Pietro Li\'o \\
Department of Computer Science \& Technology\\
University of Cambridge\\
Cambridge, United Kingdom. \\
\texttt{ben.day@cl.cam.ac.uk} \\
}

\iclrfinalcopy 
\begin{document}

\maketitle

\begin{abstract}
We propose a novel graph pooling operation using cliques as the unit pool. As this approach is purely topological, rather than featural, it is more readily interpretable, a better analogue to image coarsening than filtering or pruning techniques, and entirely nonparametric. The operation is implemented within graph convolution network (\textsc{gcn}) and Graph\textsc{sage} architectures and tested against standard graph classification benchmarks. In addition, we explore the backwards compatibility of the pooling to regular graphs, demonstrating competitive performance when replacing two-by-two pooling in standard convolutional neural networks (\textsc{cnn}s) with our mechanism.
\end{abstract}

\section{Introduction \& related work}
\label{intro}
The ongoing deep learning renaissance has proved remarkably fruitful and long lived, with state-of-the-art performance in tasks spanning the breadth of machine learning. The dominance of \textsc{cnn}s in the domain of image classification is of particular note, with superhuman performance becoming almost pedestrian. Images can be thought of as highly regular, Euclidean graphs, where pixels are nodes connected to the eight neighbouring pixels by edges. Graphs of different structures are used to represent problems from the biological, social and physical \citep{Kipf2018NeuralSystems,Gilmer2017NeuralChemistry} sciences as well as more abstract problems such as knowledge representation. Generalising the advances made for \textsc{cnn}s for use on irregular graphs has thus become an important direction for the application of deep learning, under the umbrella term geometric deep learning \citep{Bronstein2017GeometricData}.

The key operations in \textsc{cnn}s are the convolution and pooling. Convolutions extract features, with adaptations for the exploitation of locality and translational invariance. Pooling literally reduces the spatial dimensionality, aiding in the expansion of the receptive field and building consensus and saliency through coarsening. The convolution has been well adapted to non-Euclidean graphs  \citep{Kipf2016Semi-SupervisedNetworks,Defferrard2016ConvolutionalFiltering,Velickovic2017GraphNetworks,Gilmer2017NeuralChemistry} with many variations as a result of a strong interest from the research community. Pooling has not received the same treatment, unsurprisingly given it’s \textsc{cnn} and not \textsc{pnn}, and prior to recent developments the standard method was to pool globally\footnote{Take all node features and form a single set of features through some aggregation, typically element-wise max or averaging.} and then feed this into a multilayer perceptron (\textsc{mlp}) \citep{Duvenaud2015ConvolutionalFingerprints}. The current state-of-the-art methods implement gradual coarsening in hierarchies of representations \citep{gao2018graph,Cangea2018TowardsClassifiers,Ying2018HierarchicalPooling} but require hyperparameterisation, in the form of a preset allowed number of clusters or number of nodes to prune, and are not purely topologically derived. This means they are poor analogues of pooling in \textsc{cnn}s. If the aim of the community is to produce robust, transferable algorithms able to be used with many graphs then the inapplicability of these methods to regular graphs should be viewed as a serious deficiency.

In this work we introduce an operation that is purely topological, static, nonparametric and, has a natural correspondence in regular graphs and when substituted into \textsc{gnn} or traditional \textsc{cnn} for image classification, achieves performance competitive with the state-of-the-art parametric alternatives and improves on the performance of other nonparametric approaches. Being nonparametric renders the approach far more interpretable as there are no learned parameters to pick apart -- we can state quite clearly what our approach does, it puts nodes into groups where every member is connected to every other member. In addition to these properties the operation is biased towards producing a dendritic pooling hierarchy which, in combination with being static and precomputable, permits concurrent processing of the graph without loss of accuracy.

\begin{figure*}[ht]
\begin{center}

\tikzset{every picture/.style={line width=0.75pt}} 

\begin{tikzpicture}[x=0.67pt,y=0.67pt,yscale=-1,xscale=1]

\draw   (57.7,101) .. controls (57.59,95.48) and (61.98,91) .. (67.5,91) .. controls (73.02,91) and (77.59,95.48) .. (77.7,101) .. controls (77.81,106.52) and (73.42,111) .. (67.9,111) .. controls (62.38,111) and (57.81,106.52) .. (57.7,101) -- cycle ;
\draw   (83.7,116) .. controls (83.59,110.48) and (87.98,106) .. (93.5,106) .. controls (99.02,106) and (103.59,110.48) .. (103.7,116) .. controls (103.81,121.52) and (99.42,126) .. (93.9,126) .. controls (88.38,126) and (83.81,121.52) .. (83.7,116) -- cycle ;
\draw   (53.7,141) .. controls (53.59,135.48) and (57.98,131) .. (63.5,131) .. controls (69.02,131) and (73.59,135.48) .. (73.7,141) .. controls (73.81,146.52) and (69.42,151) .. (63.9,151) .. controls (58.38,151) and (53.81,146.52) .. (53.7,141) -- cycle ;
\draw   (149.7,121) .. controls (149.59,115.48) and (153.98,111) .. (159.5,111) .. controls (165.02,111) and (169.59,115.48) .. (169.7,121) .. controls (169.81,126.52) and (165.42,131) .. (159.9,131) .. controls (154.38,131) and (149.81,126.52) .. (149.7,121) -- cycle ;
\draw  [fill={rgb, 255:red, 255; green, 255; blue, 255 }  ,fill opacity=1 ] (166.7,153) .. controls (166.59,147.48) and (170.98,143) .. (176.5,143) .. controls (182.02,143) and (186.59,147.48) .. (186.7,153) .. controls (186.81,158.52) and (182.42,163) .. (176.9,163) .. controls (171.38,163) and (166.81,158.52) .. (166.7,153) -- cycle ;
\draw   (144.7,184) .. controls (144.59,178.48) and (148.98,174) .. (154.5,174) .. controls (160.02,174) and (164.59,178.48) .. (164.7,184) .. controls (164.81,189.52) and (160.42,194) .. (154.9,194) .. controls (149.38,194) and (144.81,189.52) .. (144.7,184) -- cycle ;
\draw    (176.9,163) -- (154.5,174) ;

\draw    (159.9,131) -- (176.5,143) ;

\draw    (67.9,111) -- (63.5,131) ;

\draw    (159.9,131) -- (154.5,174) ;

\draw   (130.7,144) .. controls (130.59,138.48) and (134.98,134) .. (140.5,134) .. controls (146.02,134) and (150.59,138.48) .. (150.7,144) .. controls (150.81,149.52) and (146.42,154) .. (140.9,154) .. controls (135.38,154) and (130.81,149.52) .. (130.7,144) -- cycle ;
\draw    (150.7,144) -- (166.7,153) ;

\draw    (140.9,154) -- (154.5,174) ;

\draw    (140.5,134) -- (159.9,131) ;

\draw   (98.7,215) .. controls (98.59,209.48) and (102.98,205) .. (108.5,205) .. controls (114.02,205) and (118.59,209.48) .. (118.7,215) .. controls (118.81,220.52) and (114.42,225) .. (108.9,225) .. controls (103.38,225) and (98.81,220.52) .. (98.7,215) -- cycle ;
\draw    (154.9,194) -- (118.7,215) ;

\draw [color={rgb, 255:red, 126; green, 211; blue, 33 }  ,draw opacity=1 ] [dash pattern={on 0.84pt off 2.51pt}]  (186.7,153) -- (260.57,132.53) ;
\draw [shift={(262.5,132)}, rotate = 524.52] [color={rgb, 255:red, 126; green, 211; blue, 33 }  ,draw opacity=1 ][line width=0.75]    (10.93,-3.29) .. controls (6.95,-1.4) and (3.31,-0.3) .. (0,0) .. controls (3.31,0.3) and (6.95,1.4) .. (10.93,3.29)   ;

\draw [color={rgb, 255:red, 126; green, 211; blue, 33 }  ,draw opacity=1 ] [dash pattern={on 0.84pt off 2.51pt}]  (169.7,121) -- (260.51,131.76) ;
\draw [shift={(262.5,132)}, rotate = 186.76] [color={rgb, 255:red, 126; green, 211; blue, 33 }  ,draw opacity=1 ][line width=0.75]    (10.93,-3.29) .. controls (6.95,-1.4) and (3.31,-0.3) .. (0,0) .. controls (3.31,0.3) and (6.95,1.4) .. (10.93,3.29)   ;

\draw [color={rgb, 255:red, 126; green, 211; blue, 33 }  ,draw opacity=1 ][fill={rgb, 255:red, 126; green, 211; blue, 33 }  ,fill opacity=1 ] [dash pattern={on 0.84pt off 2.51pt}]  (164.7,184) -- (260.73,132.94) ;
\draw [shift={(262.5,132)}, rotate = 512] [color={rgb, 255:red, 126; green, 211; blue, 33 }  ,draw opacity=1 ][line width=0.75]    (10.93,-3.29) .. controls (6.95,-1.4) and (3.31,-0.3) .. (0,0) .. controls (3.31,0.3) and (6.95,1.4) .. (10.93,3.29)   ;

\draw [color={rgb, 255:red, 126; green, 211; blue, 33 }  ,draw opacity=1 ] [dash pattern={on 0.84pt off 2.51pt}]  (150.7,144) -- (260.51,132.21) ;
\draw [shift={(262.5,132)}, rotate = 533.87] [color={rgb, 255:red, 126; green, 211; blue, 33 }  ,draw opacity=1 ][line width=0.75]    (10.93,-3.29) .. controls (6.95,-1.4) and (3.31,-0.3) .. (0,0) .. controls (3.31,0.3) and (6.95,1.4) .. (10.93,3.29)   ;

\draw [color={rgb, 255:red, 189; green, 16; blue, 224 }  ,draw opacity=1 ] [dash pattern={on 0.84pt off 2.51pt}]  (118.7,215) -- (286.5,203.14) ;
\draw [shift={(288.5,203)}, rotate = 535.96] [color={rgb, 255:red, 189; green, 16; blue, 224 }  ,draw opacity=1 ][line width=0.75]    (10.93,-3.29) .. controls (6.95,-1.4) and (3.31,-0.3) .. (0,0) .. controls (3.31,0.3) and (6.95,1.4) .. (10.93,3.29)   ;

\draw    (63.9,151) -- (108.5,205) ;

\draw [color={rgb, 255:red, 189; green, 16; blue, 224 }  ,draw opacity=1 ] [dash pattern={on 0.84pt off 2.51pt}]  (73.7,141) -- (286.58,202.45) ;
\draw [shift={(288.5,203)}, rotate = 196.1] [color={rgb, 255:red, 189; green, 16; blue, 224 }  ,draw opacity=1 ][line width=0.75]    (10.93,-3.29) .. controls (6.95,-1.4) and (3.31,-0.3) .. (0,0) .. controls (3.31,0.3) and (6.95,1.4) .. (10.93,3.29)   ;

\draw [color={rgb, 255:red, 74; green, 144; blue, 226 }  ,draw opacity=1 ] [dash pattern={on 0.84pt off 2.51pt}]  (73.7,141) -- (268.56,91.49) ;
\draw [shift={(270.5,91)}, rotate = 525.74] [color={rgb, 255:red, 74; green, 144; blue, 226 }  ,draw opacity=1 ][line width=0.75]    (10.93,-3.29) .. controls (6.95,-1.4) and (3.31,-0.3) .. (0,0) .. controls (3.31,0.3) and (6.95,1.4) .. (10.93,3.29)   ;

\draw [color={rgb, 255:red, 74; green, 144; blue, 226 }  ,draw opacity=1 ] [dash pattern={on 0.84pt off 2.51pt}]  (103.7,116) -- (268.52,91.3) ;
\draw [shift={(270.5,91)}, rotate = 531.48] [color={rgb, 255:red, 74; green, 144; blue, 226 }  ,draw opacity=1 ][line width=0.75]    (10.93,-3.29) .. controls (6.95,-1.4) and (3.31,-0.3) .. (0,0) .. controls (3.31,0.3) and (6.95,1.4) .. (10.93,3.29)   ;

\draw [color={rgb, 255:red, 74; green, 144; blue, 226 }  ,draw opacity=1 ] [dash pattern={on 0.84pt off 2.51pt}]  (77.7,101) -- (268.5,91.1) ;
\draw [shift={(270.5,91)}, rotate = 537.03] [color={rgb, 255:red, 74; green, 144; blue, 226 }  ,draw opacity=1 ][line width=0.75]    (10.93,-3.29) .. controls (6.95,-1.4) and (3.31,-0.3) .. (0,0) .. controls (3.31,0.3) and (6.95,1.4) .. (10.93,3.29)   ;

\draw   (57.7,64) .. controls (57.59,58.48) and (61.98,54) .. (67.5,54) .. controls (73.02,54) and (77.59,58.48) .. (77.7,64) .. controls (77.81,69.52) and (73.42,74) .. (67.9,74) .. controls (62.38,74) and (57.81,69.52) .. (57.7,64) -- cycle ;
\draw    (67.9,74) -- (67.5,91) ;

\draw [color={rgb, 255:red, 208; green, 2; blue, 27 }  ,draw opacity=1 ] [dash pattern={on 0.84pt off 2.51pt}]  (77.7,64) -- (321.5,76.89) ;
\draw [shift={(323.5,77)}, rotate = 183.03] [color={rgb, 255:red, 208; green, 2; blue, 27 }  ,draw opacity=1 ][line width=0.75]    (10.93,-3.29) .. controls (6.95,-1.4) and (3.31,-0.3) .. (0,0) .. controls (3.31,0.3) and (6.95,1.4) .. (10.93,3.29)   ;

\draw  [fill={rgb, 255:red, 208; green, 2; blue, 27 }  ,fill opacity=1 ] (319.5,77) .. controls (319.39,71.48) and (323.78,67) .. (329.3,67) .. controls (334.82,67) and (339.39,71.48) .. (339.5,77) .. controls (339.61,82.52) and (335.22,87) .. (329.7,87) .. controls (324.18,87) and (319.61,82.52) .. (319.5,77) -- cycle ;
\draw  [fill={rgb, 255:red, 74; green, 144; blue, 226 }  ,fill opacity=1 ] (270.5,91) .. controls (270.39,85.48) and (274.78,81) .. (280.3,81) .. controls (285.82,81) and (290.39,85.48) .. (290.5,91) .. controls (290.61,96.52) and (286.22,101) .. (280.7,101) .. controls (275.18,101) and (270.61,96.52) .. (270.5,91) -- cycle ;
\draw  [fill={rgb, 255:red, 126; green, 211; blue, 33 }  ,fill opacity=1 ] (262.5,132) .. controls (262.39,126.48) and (266.78,122) .. (272.3,122) .. controls (277.82,122) and (282.39,126.48) .. (282.5,132) .. controls (282.61,137.52) and (278.22,142) .. (272.7,142) .. controls (267.18,142) and (262.61,137.52) .. (262.5,132) -- cycle ;
\draw  [fill={rgb, 255:red, 189; green, 16; blue, 224 }  ,fill opacity=1 ] (288.5,203) .. controls (288.39,197.48) and (292.78,193) .. (298.3,193) .. controls (303.82,193) and (308.39,197.48) .. (308.5,203) .. controls (308.61,208.52) and (304.22,213) .. (298.7,213) .. controls (293.18,213) and (288.61,208.52) .. (288.5,203) -- cycle ;
\draw    (319.5,77) -- (290.5,91) ;

\draw    (303,195) -- (272.7,142) ;

\draw    (280.7,101) -- (272.3,122) ;

\draw    (280.7,101) -- (303,195) ;

\draw [color={rgb, 255:red, 245; green, 166; blue, 35 }  ,draw opacity=1 ] [dash pattern={on 0.84pt off 2.51pt}]  (308.5,203) -- (423.64,156.75) ;
\draw [shift={(425.5,156)}, rotate = 518.11] [color={rgb, 255:red, 245; green, 166; blue, 35 }  ,draw opacity=1 ][line width=0.75]    (10.93,-3.29) .. controls (6.95,-1.4) and (3.31,-0.3) .. (0,0) .. controls (3.31,0.3) and (6.95,1.4) .. (10.93,3.29)   ;

\draw [color={rgb, 255:red, 245; green, 166; blue, 35 }  ,draw opacity=1 ] [dash pattern={on 0.84pt off 2.51pt}]  (282.5,132) -- (423.53,155.67) ;
\draw [shift={(425.5,156)}, rotate = 189.53] [color={rgb, 255:red, 245; green, 166; blue, 35 }  ,draw opacity=1 ][line width=0.75]    (10.93,-3.29) .. controls (6.95,-1.4) and (3.31,-0.3) .. (0,0) .. controls (3.31,0.3) and (6.95,1.4) .. (10.93,3.29)   ;

\draw [color={rgb, 255:red, 245; green, 166; blue, 35 }  ,draw opacity=1 ] [dash pattern={on 0.84pt off 2.51pt}]  (290.5,91) -- (423.7,155.13) ;
\draw [shift={(425.5,156)}, rotate = 205.71] [color={rgb, 255:red, 245; green, 166; blue, 35 }  ,draw opacity=1 ][line width=0.75]    (10.93,-3.29) .. controls (6.95,-1.4) and (3.31,-0.3) .. (0,0) .. controls (3.31,0.3) and (6.95,1.4) .. (10.93,3.29)   ;

\draw [color={rgb, 255:red, 80; green, 227; blue, 194 }  ,draw opacity=1 ] [dash pattern={on 0.84pt off 2.51pt}]  (339.5,77) -- (426.61,107.34) ;
\draw [shift={(428.5,108)}, rotate = 199.2] [color={rgb, 255:red, 80; green, 227; blue, 194 }  ,draw opacity=1 ][line width=0.75]    (10.93,-3.29) .. controls (6.95,-1.4) and (3.31,-0.3) .. (0,0) .. controls (3.31,0.3) and (6.95,1.4) .. (10.93,3.29)   ;

\draw  [fill={rgb, 255:red, 80; green, 227; blue, 194 }  ,fill opacity=1 ] (428.5,108) .. controls (428.39,102.48) and (432.78,98) .. (438.3,98) .. controls (443.82,98) and (448.39,102.48) .. (448.5,108) .. controls (448.61,113.52) and (444.22,118) .. (438.7,118) .. controls (433.18,118) and (428.61,113.52) .. (428.5,108) -- cycle ;
\draw  [fill={rgb, 255:red, 245; green, 166; blue, 35 }  ,fill opacity=1 ] (425.5,156) .. controls (425.39,150.48) and (429.78,146) .. (435.3,146) .. controls (440.82,146) and (445.39,150.48) .. (445.5,156) .. controls (445.61,161.52) and (441.22,166) .. (435.7,166) .. controls (430.18,166) and (425.61,161.52) .. (425.5,156) -- cycle ;
\draw    (438.7,118) -- (435.3,146) ;

\draw  [dash pattern={on 0.84pt off 2.51pt}]  (448.5,108) -- (545.63,144.3) ;
\draw [shift={(547.5,145)}, rotate = 200.49] [color={rgb, 255:red, 0; green, 0; blue, 0 }  ][line width=0.75]    (10.93,-3.29) .. controls (6.95,-1.4) and (3.31,-0.3) .. (0,0) .. controls (3.31,0.3) and (6.95,1.4) .. (10.93,3.29)   ;

\draw  [dash pattern={on 0.84pt off 2.51pt}]  (445.5,156) -- (545.51,145.21) ;
\draw [shift={(547.5,145)}, rotate = 533.8399999999999] [color={rgb, 255:red, 0; green, 0; blue, 0 }  ][line width=0.75]    (10.93,-3.29) .. controls (6.95,-1.4) and (3.31,-0.3) .. (0,0) .. controls (3.31,0.3) and (6.95,1.4) .. (10.93,3.29)   ;

\draw  [color={rgb, 255:red, 0; green, 0; blue, 0 }  ,draw opacity=1 ][fill={rgb, 255:red, 0; green, 0; blue, 0 }  ,fill opacity=1 ] (547.5,145) .. controls (547.39,139.48) and (551.78,135) .. (557.3,135) .. controls (562.82,135) and (567.39,139.48) .. (567.5,145) .. controls (567.61,150.52) and (563.22,155) .. (557.7,155) .. controls (552.18,155) and (547.61,150.52) .. (547.5,145) -- cycle ;
\draw  [color={rgb, 255:red, 126; green, 211; blue, 33 }  ,draw opacity=1 ][line width=1.5]  (136,120) .. controls (154.5,93) and (205.5,116) .. (195.5,148) .. controls (185.5,180) and (144.5,229) .. (136,180) .. controls (127.5,131) and (117.5,147) .. (136,120) -- cycle ;
\draw  [color={rgb, 255:red, 189; green, 16; blue, 224 }  ,draw opacity=1 ][line width=1.5]  (67.2,122) .. controls (86.5,135) and (87.5,147) .. (99.5,174) .. controls (111.5,201) and (119.5,208) .. (131.5,223) .. controls (143.5,238) and (88.5,236) .. (60.5,214) .. controls (32.5,192) and (47.9,109) .. (67.2,122) -- cycle ;
\draw  [color={rgb, 255:red, 74; green, 144; blue, 226 }  ,draw opacity=1 ][line width=1.5]  (55.7,91) .. controls (75.7,81) and (95.1,89) .. (105.7,102) .. controls (116.3,115) and (101.7,123) .. (88.7,146) .. controls (75.7,169) and (69.7,180) .. (49.7,150) .. controls (29.7,120) and (35.7,101) .. (55.7,91) -- cycle ;
\draw  [color={rgb, 255:red, 208; green, 2; blue, 27 }  ,draw opacity=1 ][line width=1.5]  (65.5,47) .. controls (84.5,49) and (77.5,46) .. (83.5,88) .. controls (89.5,130) and (42.5,121) .. (48.5,99) .. controls (54.5,77) and (46.5,45) .. (65.5,47) -- cycle ;
\draw  [color={rgb, 255:red, 245; green, 166; blue, 35 }  ,draw opacity=1 ][line width=1.5]  (263.5,78) .. controls (270.5,64) and (306.5,60) .. (305,84) .. controls (303.5,108) and (301.5,120) .. (315.5,179) .. controls (329.5,238) and (324.5,243) .. (284.5,223) .. controls (244.5,203) and (256.5,92) .. (263.5,78) -- cycle ;
\draw  [color={rgb, 255:red, 80; green, 227; blue, 194 }  ,draw opacity=1 ][line width=1.5]  (310.5,63) .. controls (330.5,53) and (328.5,58) .. (330.5,58) .. controls (332.5,58) and (378.5,86) .. (347.5,89) .. controls (316.5,92) and (288.5,135) .. (268.5,105) .. controls (248.5,75) and (290.5,73) .. (310.5,63) -- cycle ;
\draw  [line width=1.5]  (426.5,89) .. controls (446.5,79) and (462.5,91) .. (461.5,105) .. controls (460.5,119) and (447.5,128) .. (461.5,159) .. controls (475.5,190) and (400.5,187) .. (411,161) .. controls (421.5,135) and (406.5,99) .. (426.5,89) -- cycle ;
\draw   (58.7,198) .. controls (58.59,192.48) and (62.98,188) .. (68.5,188) .. controls (74.02,188) and (78.59,192.48) .. (78.7,198) .. controls (78.81,203.52) and (74.42,208) .. (68.9,208) .. controls (63.38,208) and (58.81,203.52) .. (58.7,198) -- cycle ;
\draw    (68.9,208) -- (98.7,215) ;

\draw    (63.9,151) -- (68.5,188) ;

\draw [color={rgb, 255:red, 144; green, 19; blue, 254 }  ,draw opacity=1 ] [dash pattern={on 0.84pt off 2.51pt}]  (78.7,198) -- (286.5,202.95) ;
\draw [shift={(288.5,203)}, rotate = 181.37] [color={rgb, 255:red, 144; green, 19; blue, 254 }  ,draw opacity=1 ][line width=0.75]    (10.93,-3.29) .. controls (6.95,-1.4) and (3.31,-0.3) .. (0,0) .. controls (3.31,0.3) and (6.95,1.4) .. (10.93,3.29)   ;

\draw    (77.7,101) -- (93.5,106) ;

\draw    (93.9,126) -- (73.7,141) ;

\draw    (103.7,116) -- (149.7,121) ;

\draw  [color={rgb, 255:red, 155; green, 155; blue, 155 }  ,draw opacity=1 ][dash pattern={on 5.63pt off 4.5pt}][line width=1.5]  (92.5,99) .. controls (107.5,93) and (135.5,101) .. (158.5,103) .. controls (181.5,105) and (176.5,147) .. (160.5,138) .. controls (144.5,129) and (128.5,127) .. (101.5,131) .. controls (74.5,135) and (77.5,105) .. (92.5,99) -- cycle ;
\draw  [color={rgb, 255:red, 139; green, 87; blue, 42 }  ,draw opacity=1 ][dash pattern={on 5.63pt off 4.5pt}][line width=1.5]  (109.5,197) .. controls (131.5,199) and (155.7,149.5) .. (165.7,168.5) .. controls (175.7,187.5) and (177.5,195) .. (159.5,203) .. controls (141.5,211) and (141.5,212) .. (116.5,227) .. controls (91.5,242) and (87.5,195) .. (109.5,197) -- cycle ;

\draw (116,16) node  [align=left] {Original Graph};
\draw (295,21) node  [align=left] {Graph after \\1 pooling layer};
\draw (450,22) node  [align=left] {Graph after\\2 pooling layers};
\draw (575,23) node  [align=left] {Graph after\\3 pooling layers};
\draw (55,45) node  [align=left] {1};
\draw (33,121) node  [align=left] {2};
\draw (41,193) node  [align=left] {3};
\draw (148,216) node  [align=left] {4};
\draw (193,116) node  [align=left] {5};
\draw (134,94) node  [align=left] {6};
\draw (329.5,77) node  [align=left] {1};
\draw (280.5,91) node  [align=left] {2};
\draw (298.5,203) node  [align=left] {3};
\draw (272.5,132) node  [align=left] {5};
\draw (349,58) node  [align=left] {A};
\draw (331,218) node  [align=left] {B};
\draw (438.5,108) node  [align=left] {A};
\draw (435.5,156) node  [align=left] {B};
\draw (463.5,86) node  [align=left] {J};
\draw (557.5,145) node [color={rgb, 255:red, 255; green, 255; blue, 255 }  ,opacity=1 ] [align=left] {J};

\end{tikzpicture}

\caption{Clique Pooling for an irregular graph. The colored borders represent the maximal cliques (also labeled with numbers or letters next to them), dotted arrows indicate the cliques to which the nodes are assigned. Notice that although some of the nodes belong to more than one clique they do not necessarily contribute to the respective node in the coarsened graph. This is the case for the node belonging to the red (1) and blue (2) maximal clique. Since the blue clique is bigger (in terms of nodes), the node is assigned to the blue (2) cluster only. In the case of the node intersecting the blue (2) and purple (3) maximal cliques, the node is assigned to both cliques since the cliques have the same size. Hence, it contributes to the features of both of the respective nodes in the coarsened graph. The grey maximal clique (6) is not represented in the new coarsened graph since the nodes in that clique have already been assigned to larger cliques: the green (5) and blue (2) cliques. The nodes is the coarsened graph are connected if any two nodes in the respective cliques are connected. }
\label{icml-historical}
\end{center}
\vskip -0.2in
\end{figure*}
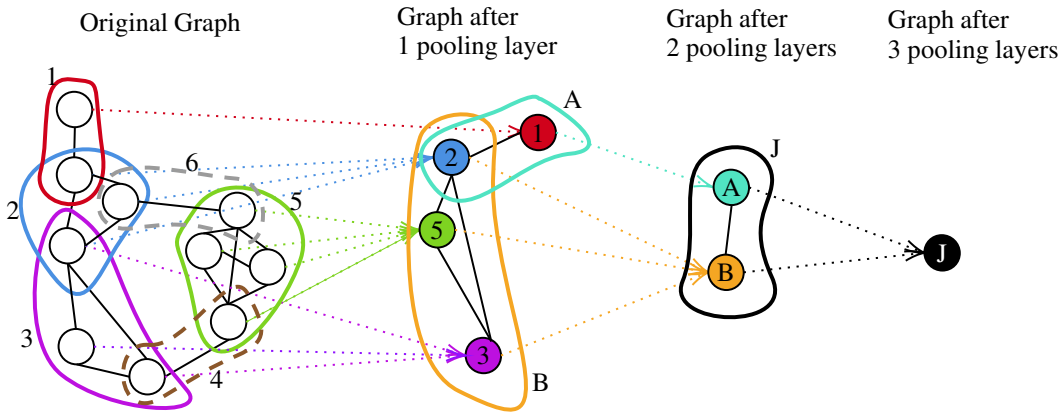

\section{Proposed Method}
\label{proposedMethod}
\subsection{Preliminaries}
We use the standard representation of graphs in graph classification tasks. The graph $G$ with $N$ nodes is represented as a pair $(A, X)$ where the adjacency matrix $A \in \mathbb{R}^{N \times N}$ and the node feature matrix, with $F$ features per node, $X \in \mathbb{R}^{N \times F}$. Additionally, our method assumes undirected graphs.

\paragraph{Graph Convolution} In principle, our method does not require a convolution operator. However, a \textsc{cnn}-like architecture requires one. The graph convolution needs to be inductive. In our experiments we use Graph\textsc{sage} and \textsc{gcn} as particular implementations of the message-passing scheme.

\paragraph{Readout Function}
We use readout functions to combine pools of nodes to form the features in the coarsened graph and to generate representations of entire layers to be fed into the \textsc{mlp} classifier. For pool aggregation we take the mean over nodes in the pool. For whole-layer representation we use the concatenation of the mean and maximum over all nodes in the layer. The whole input to the \textsc{mlp} is the concatenation of each layer representation. \textsc{Cnn}s typically take the maximum within pools until the final layer when a whole-field pool reduces the 3-tensor, $[H,W,F]$, to a vector, $[1,1,F]\rightarrow[F]$.

\subsection{Proposed Operation}

Our approach is to coarsen the graph by aggregating maximal cliques. We  attempt to limit the dispersal of nodes by assigning each to a single clique-pool, ranked by size, where possible. Only in the case of equally large options is the node assigned to multiple pools.

To accomplish this, the maximal cliques are found using the Bron-Kerbosch algorithm \citep{Bron1973AlgorithmGraph} with modifications shown to improve performance on large real-world graphs \citep{Eppstein2010ListingTime,Eppstein2011ListingGraphs}. Nodes are assigned to pools greedily starting with the largest. When a node has been assigned it is removed from the remaining smaller pools and does not count towards that pool's size for the assignment of other nodes. Edges are inherited from the clique members such that if a node in one clique shares an edge with a node in another, the cliques will share an edge in the coarsened graph. The resulting graph shares some features with the clique-graph\footnote{The graph formed by maximal cliques where edges occur where cliques intersect.} though with fewer nodes and more edges in all but the simplest cases. The assigned cliques are then pooled using whichever pooling function is desired, average- and max-pooling in the experiments presented here.

\begin{remark}[Convergence]
Given a connected and finite graph, the clique-pooling operator converges to a single node after finitely many steps.
\end{remark}

The proof for the above can be found in Appendix \ref{appenC}. This guarantees that in any graph we can have a \textsc{cnn}-like architecture. As the approach is based only on the topology of the original graph it is both nonparametric and static. This means that the pools can be precomputed and opens the door to greater parallelization for dealing with very large graphs. Having found the entire pooling structure it is trivial to generate the dependency diagram, allowing graph partitions to be loaded and operated on separately, which will be key to dealing with very-large graphs, as discussed in \citealt{Zhang2005Genome-scaleBiology} for the case of systems biology.

\subsection{Application to Images as Regular Graphs}
\label{applicationImages}
\begin{wrapfigure}{r}{6cm}
\begin{center}
\tikzset{every picture/.style={line width=0.75pt}} 
\begin{tikzpicture}[x=0.75pt,y=0.75pt,yscale=-0.9,xscale=0.9]
\draw    (45.98,45.94) -- (64.06,64.06) ;
\draw    (64.06,45.94) -- (45.98,64.06) ;
\draw    (94.02,45.94) -- (75.94,64.06) ;
\draw    (124.2,45.94) -- (106.12,64.06) ;
\draw    (64.06,76.08) -- (45.98,94.2) ;
\draw    (94.06,75.94) -- (75.98,94.06) ;
\draw    (124.2,75.94) -- (106.12,94.06) ;
\draw    (124.2,106.08) -- (106.12,124.2) ;
\draw    (94.06,105.94) -- (75.98,124.06) ;
\draw    (64.06,106.08) -- (45.98,124.2) ;
\draw    (75.98,45.94) -- (94.06,64.06) ;
\draw    (105.94,45.94) -- (124.02,64.06) ;
\draw    (45.98,75.94) -- (64.06,94.06) ;
\draw    (75.98,75.94) -- (94.06,94.06) ;
\draw    (106.12,76.08) -- (124.2,94.2) ;
\draw    (45.98,106.08) -- (64.06,124.2) ;
\draw    (75.98,106.08) -- (94.06,124.2) ;
\draw    (106.12,106.08) -- (124.2,124.2) ;
\draw   (32.16,40) .. controls (32.16,35.65) and (35.69,32.13) .. (40.04,32.13) .. controls (44.38,32.13) and (47.91,35.65) .. (47.91,40) .. controls (47.91,44.35) and (44.38,47.88) .. (40.04,47.88) .. controls (35.69,47.88) and (32.16,44.35) .. (32.16,40) -- cycle ;
\draw   (62.13,40) .. controls (62.13,35.65) and (65.65,32.13) .. (70,32.13) .. controls (74.35,32.13) and (77.88,35.65) .. (77.88,40) .. controls (77.88,44.35) and (74.35,47.88) .. (70,47.88) .. controls (65.65,47.88) and (62.13,44.35) .. (62.13,40) -- cycle ;
\draw   (92.13,40) .. controls (92.13,35.65) and (95.65,32.13) .. (100,32.13) .. controls (104.35,32.13) and (107.88,35.65) .. (107.88,40) .. controls (107.88,44.35) and (104.35,47.88) .. (100,47.88) .. controls (95.65,47.88) and (92.13,44.35) .. (92.13,40) -- cycle ;
\draw   (122.27,40) .. controls (122.27,35.65) and (125.79,32.13) .. (130.14,32.13) .. controls (134.49,32.13) and (138.02,35.65) .. (138.02,40) .. controls (138.02,44.35) and (134.49,47.88) .. (130.14,47.88) .. controls (125.79,47.88) and (122.27,44.35) .. (122.27,40) -- cycle ;
\draw   (32.16,70) .. controls (32.16,65.65) and (35.69,62.13) .. (40.04,62.13) .. controls (44.38,62.13) and (47.91,65.65) .. (47.91,70) .. controls (47.91,74.35) and (44.38,77.88) .. (40.04,77.88) .. controls (35.69,77.88) and (32.16,74.35) .. (32.16,70) -- cycle ;
\draw   (62.13,70) .. controls (62.13,65.65) and (65.65,62.13) .. (70,62.13) .. controls (74.35,62.13) and (77.88,65.65) .. (77.88,70) .. controls (77.88,74.35) and (74.35,77.88) .. (70,77.88) .. controls (65.65,77.88) and (62.13,74.35) .. (62.13,70) -- cycle ;
\draw   (92.13,70) .. controls (92.13,65.65) and (95.65,62.13) .. (100,62.13) .. controls (104.35,62.13) and (107.88,65.65) .. (107.88,70) .. controls (107.88,74.35) and (104.35,77.88) .. (100,77.88) .. controls (95.65,77.88) and (92.13,74.35) .. (92.13,70) -- cycle ;
\draw   (122.27,70) .. controls (122.27,65.65) and (125.79,62.13) .. (130.14,62.13) .. controls (134.49,62.13) and (138.02,65.65) .. (138.02,70) .. controls (138.02,74.35) and (134.49,77.88) .. (130.14,77.88) .. controls (125.79,77.88) and (122.27,74.35) .. (122.27,70) -- cycle ;
\draw   (32.16,100.14) .. controls (32.16,95.79) and (35.69,92.27) .. (40.04,92.27) .. controls (44.38,92.27) and (47.91,95.79) .. (47.91,100.14) .. controls (47.91,104.49) and (44.38,108.02) .. (40.04,108.02) .. controls (35.69,108.02) and (32.16,104.49) .. (32.16,100.14) -- cycle ;
\draw   (62.13,100.14) .. controls (62.13,95.79) and (65.65,92.27) .. (70,92.27) .. controls (74.35,92.27) and (77.88,95.79) .. (77.88,100.14) .. controls (77.88,104.49) and (74.35,108.02) .. (70,108.02) .. controls (65.65,108.02) and (62.13,104.49) .. (62.13,100.14) -- cycle ;
\draw   (92.13,100.14) .. controls (92.13,95.79) and (95.65,92.27) .. (100,92.27) .. controls (104.35,92.27) and (107.88,95.79) .. (107.88,100.14) .. controls (107.88,104.49) and (104.35,108.02) .. (100,108.02) .. controls (95.65,108.02) and (92.13,104.49) .. (92.13,100.14) -- cycle ;
\draw   (122.27,100.14) .. controls (122.27,95.79) and (125.79,92.27) .. (130.14,92.27) .. controls (134.49,92.27) and (138.02,95.79) .. (138.02,100.14) .. controls (138.02,104.49) and (134.49,108.02) .. (130.14,108.02) .. controls (125.79,108.02) and (122.27,104.49) .. (122.27,100.14) -- cycle ;
\draw   (32.16,130.14) .. controls (32.16,125.79) and (35.69,122.27) .. (40.04,122.27) .. controls (44.38,122.27) and (47.91,125.79) .. (47.91,130.14) .. controls (47.91,134.49) and (44.38,138.02) .. (40.04,138.02) .. controls (35.69,138.02) and (32.16,134.49) .. (32.16,130.14) -- cycle ;
\draw   (62.13,130.14) .. controls (62.13,125.79) and (65.65,122.27) .. (70,122.27) .. controls (74.35,122.27) and (77.88,125.79) .. (77.88,130.14) .. controls (77.88,134.49) and (74.35,138.02) .. (70,138.02) .. controls (65.65,138.02) and (62.13,134.49) .. (62.13,130.14) -- cycle ;
\draw   (92.13,130.14) .. controls (92.13,125.79) and (95.65,122.27) .. (100,122.27) .. controls (104.35,122.27) and (107.88,125.79) .. (107.88,130.14) .. controls (107.88,134.49) and (104.35,138.02) .. (100,138.02) .. controls (95.65,138.02) and (92.13,134.49) .. (92.13,130.14) -- cycle ;
\draw   (122.27,130.14) .. controls (122.27,125.79) and (125.79,122.27) .. (130.14,122.27) .. controls (134.49,122.27) and (138.02,125.79) .. (138.02,130.14) .. controls (138.02,134.49) and (134.49,138.02) .. (130.14,138.02) .. controls (125.79,138.02) and (122.27,134.49) .. (122.27,130.14) -- cycle ;
\draw    (40.04,47.88) -- (40.04,62.13) ;
\draw    (40.04,77.88) -- (40.04,92.27) ;
\draw    (40.04,108.02) -- (40.04,122.27) ;
\draw    (70,47.88) -- (70,62.13) ;
\draw    (70,77.88) -- (70,92.27) ;
\draw    (70,108.02) -- (70,122.27) ;
\draw    (100,47.88) -- (100,62.13) ;
\draw    (100,77.88) -- (100,92.27) ;
\draw    (100,108.02) -- (100,122.27) ;
\draw    (130.14,47.88) -- (130.14,62.13) ;
\draw    (130.14,77.88) -- (130.14,92.27) ;
\draw    (130.14,108.02) -- (130.14,122.27) ;
\draw    (47.91,40) -- (62.13,40) ;
\draw    (77.88,40) -- (92.13,40) ;
\draw    (107.88,40) -- (122.27,40) ;
\draw    (62.13,70) -- (47.91,70) ;
\draw    (92.13,70) -- (77.88,70) ;
\draw    (122.27,70) -- (107.88,70) ;
\draw    (62.13,100.14) -- (47.91,100.14) ;
\draw    (92.13,100.14) -- (77.88,100.14) ;
\draw    (122.27,100.14) -- (107.88,100.14) ;
\draw    (122.27,130.14) -- (107.88,130.14) ;
\draw    (92.13,130.14) -- (77.88,130.14) ;
\draw    (62.13,130.14) -- (47.91,130.14) ;
\draw  [color={rgb, 255:red, 40; green, 127; blue, 229 }  ,draw opacity=1 ] (42.18,42.56) -- (67.86,42.56) -- (67.86,68.24) -- (42.18,68.24) -- cycle ;
\draw  [color={rgb, 255:red, 40; green, 127; blue, 229 }  ,draw opacity=1 ] (72.18,42.56) -- (97.86,42.56) -- (97.86,68.24) -- (72.18,68.24) -- cycle ;
\draw  [color={rgb, 255:red, 40; green, 127; blue, 229 }  ,draw opacity=1 ] (102.32,42.56) -- (128,42.56) -- (128,68.24) -- (102.32,68.24) -- cycle ;
\draw  [color={rgb, 255:red, 40; green, 127; blue, 229 }  ,draw opacity=1 ] (42.18,72.3) -- (67.86,72.3) -- (67.86,97.98) -- (42.18,97.98) -- cycle ;
\draw  [color={rgb, 255:red, 40; green, 127; blue, 229 }  ,draw opacity=1 ] (42.18,102.3) -- (67.86,102.3) -- (67.86,127.98) -- (42.18,127.98) -- cycle ;
\draw  [color={rgb, 255:red, 40; green, 127; blue, 229 }  ,draw opacity=1 ] (72.18,72.16) -- (97.86,72.16) -- (97.86,97.84) -- (72.18,97.84) -- cycle ;
\draw  [color={rgb, 255:red, 40; green, 127; blue, 229 }  ,draw opacity=1 ] (102.32,72.3) -- (128,72.3) -- (128,97.98) -- (102.32,97.98) -- cycle ;
\draw  [color={rgb, 255:red, 40; green, 127; blue, 229 }  ,draw opacity=1 ] (72.18,102.16) -- (97.86,102.16) -- (97.86,127.84) -- (72.18,127.84) -- cycle ;
\draw  [color={rgb, 255:red, 40; green, 127; blue, 229 }  ,draw opacity=1 ] (102.32,102.3) -- (128,102.3) -- (128,127.98) -- (102.32,127.98) -- cycle ;
\draw [color={rgb, 255:red, 40; green, 127; blue, 229 }  ,draw opacity=1 ]   (191.48,60.94) -- (209.56,79.06) ;
\draw [color={rgb, 255:red, 40; green, 127; blue, 229 }  ,draw opacity=1 ]   (209.56,60.94) -- (191.48,79.06) ;
\draw [color={rgb, 255:red, 40; green, 127; blue, 229 }  ,draw opacity=1 ]   (239.52,60.94) -- (221.44,79.06) ;
\draw [color={rgb, 255:red, 40; green, 127; blue, 229 }  ,draw opacity=1 ]   (209.56,91.08) -- (191.48,109.2) ;
\draw [color={rgb, 255:red, 40; green, 127; blue, 229 }  ,draw opacity=1 ]   (239.56,90.94) -- (221.48,109.06) ;
\draw [color={rgb, 255:red, 40; green, 127; blue, 229 }  ,draw opacity=1 ]   (221.48,60.94) -- (239.56,79.06) ;
\draw [color={rgb, 255:red, 40; green, 127; blue, 229 }  ,draw opacity=1 ]   (191.48,90.94) -- (209.56,109.06) ;
\draw [color={rgb, 255:red, 40; green, 127; blue, 229 }  ,draw opacity=1 ]   (221.48,90.94) -- (239.56,109.06) ;
\draw  [color={rgb, 255:red, 40; green, 127; blue, 229 }  ,draw opacity=1 ] (177.66,55) .. controls (177.66,50.65) and (181.19,47.13) .. (185.54,47.13) .. controls (189.88,47.13) and (193.41,50.65) .. (193.41,55) .. controls (193.41,59.35) and (189.88,62.88) .. (185.54,62.88) .. controls (181.19,62.88) and (177.66,59.35) .. (177.66,55) -- cycle ;
\draw  [color={rgb, 255:red, 40; green, 127; blue, 229 }  ,draw opacity=1 ] (207.63,55) .. controls (207.63,50.65) and (211.15,47.13) .. (215.5,47.13) .. controls (219.85,47.13) and (223.38,50.65) .. (223.38,55) .. controls (223.38,59.35) and (219.85,62.88) .. (215.5,62.88) .. controls (211.15,62.88) and (207.63,59.35) .. (207.63,55) -- cycle ;
\draw  [color={rgb, 255:red, 40; green, 127; blue, 229 }  ,draw opacity=1 ] (237.63,55) .. controls (237.63,50.65) and (241.15,47.13) .. (245.5,47.13) .. controls (249.85,47.13) and (253.37,50.65) .. (253.37,55) .. controls (253.37,59.35) and (249.85,62.88) .. (245.5,62.88) .. controls (241.15,62.88) and (237.63,59.35) .. (237.63,55) -- cycle ;
\draw  [color={rgb, 255:red, 40; green, 127; blue, 229 }  ,draw opacity=1 ] (177.66,85) .. controls (177.66,80.65) and (181.19,77.13) .. (185.54,77.13) .. controls (189.88,77.13) and (193.41,80.65) .. (193.41,85) .. controls (193.41,89.35) and (189.88,92.88) .. (185.54,92.88) .. controls (181.19,92.88) and (177.66,89.35) .. (177.66,85) -- cycle ;
\draw  [color={rgb, 255:red, 40; green, 127; blue, 229 }  ,draw opacity=1 ] (207.63,85) .. controls (207.63,80.65) and (211.15,77.13) .. (215.5,77.13) .. controls (219.85,77.13) and (223.38,80.65) .. (223.38,85) .. controls (223.38,89.35) and (219.85,92.88) .. (215.5,92.88) .. controls (211.15,92.88) and (207.63,89.35) .. (207.63,85) -- cycle ;
\draw  [color={rgb, 255:red, 40; green, 127; blue, 229 }  ,draw opacity=1 ] (237.63,85) .. controls (237.63,80.65) and (241.15,77.13) .. (245.5,77.13) .. controls (249.85,77.13) and (253.37,80.65) .. (253.37,85) .. controls (253.37,89.35) and (249.85,92.88) .. (245.5,92.88) .. controls (241.15,92.88) and (237.63,89.35) .. (237.63,85) -- cycle ;
\draw  [color={rgb, 255:red, 40; green, 127; blue, 229 }  ,draw opacity=1 ] (177.66,115.14) .. controls (177.66,110.79) and (181.19,107.27) .. (185.54,107.27) .. controls (189.88,107.27) and (193.41,110.79) .. (193.41,115.14) .. controls (193.41,119.49) and (189.88,123.02) .. (185.54,123.02) .. controls (181.19,123.02) and (177.66,119.49) .. (177.66,115.14) -- cycle ;
\draw  [color={rgb, 255:red, 40; green, 127; blue, 229 }  ,draw opacity=1 ] (207.63,115.14) .. controls (207.63,110.79) and (211.15,107.27) .. (215.5,107.27) .. controls (219.85,107.27) and (223.38,110.79) .. (223.38,115.14) .. controls (223.38,119.49) and (219.85,123.02) .. (215.5,123.02) .. controls (211.15,123.02) and (207.63,119.49) .. (207.63,115.14) -- cycle ;
\draw  [color={rgb, 255:red, 40; green, 127; blue, 229 }  ,draw opacity=1 ] (237.63,115.14) .. controls (237.63,110.79) and (241.15,107.27) .. (245.5,107.27) .. controls (249.85,107.27) and (253.37,110.79) .. (253.37,115.14) .. controls (253.37,119.49) and (249.85,123.02) .. (245.5,123.02) .. controls (241.15,123.02) and (237.63,119.49) .. (237.63,115.14) -- cycle ;
\draw [color={rgb, 255:red, 40; green, 127; blue, 229 }  ,draw opacity=1 ]   (185.54,62.88) -- (185.54,77.13) ;
\draw [color={rgb, 255:red, 40; green, 127; blue, 229 }  ,draw opacity=1 ]   (185.54,92.88) -- (185.54,107.27) ;
\draw [color={rgb, 255:red, 40; green, 127; blue, 229 }  ,draw opacity=1 ]   (215.5,62.88) -- (215.5,77.13) ;
\draw [color={rgb, 255:red, 40; green, 127; blue, 229 }  ,draw opacity=1 ]   (215.5,92.88) -- (215.5,107.27) ;
\draw [color={rgb, 255:red, 40; green, 127; blue, 229 }  ,draw opacity=1 ]   (245.5,62.88) -- (245.5,77.13) ;
\draw [color={rgb, 255:red, 40; green, 127; blue, 229 }  ,draw opacity=1 ]   (245.5,92.88) -- (245.5,107.27) ;
\draw [color={rgb, 255:red, 40; green, 127; blue, 229 }  ,draw opacity=1 ]   (193.41,55) -- (207.63,55) ;
\draw [color={rgb, 255:red, 40; green, 127; blue, 229 }  ,draw opacity=1 ]   (223.38,55) -- (237.63,55) ;
\draw [color={rgb, 255:red, 40; green, 127; blue, 229 }  ,draw opacity=1 ]   (207.63,85) -- (193.41,85) ;
\draw [color={rgb, 255:red, 40; green, 127; blue, 229 }  ,draw opacity=1 ]   (237.63,85) -- (223.38,85) ;
\draw [color={rgb, 255:red, 40; green, 127; blue, 229 }  ,draw opacity=1 ]   (207.63,115.14) -- (193.41,115.14) ;
\draw [color={rgb, 255:red, 40; green, 127; blue, 229 }  ,draw opacity=1 ]   (237.63,115.14) -- (223.38,115.14) ;
\draw    (137,85) -- (159.38,85) ;
\draw [shift={(161.38,85)}, rotate = 180] [color={rgb, 255:red, 0; green, 0; blue, 0 }  ][line width=0.75]    (10.93,-3.29) .. controls (6.95,-1.4) and (3.31,-0.3) .. (0,0) .. controls (3.31,0.3) and (6.95,1.4) .. (10.93,3.29)   ;
\draw [color={rgb, 255:red, 40; green, 127; blue, 229 }  ,draw opacity=1 ]   (191.48,60.94) -- (239.56,79.06) ;
\draw [color={rgb, 255:red, 40; green, 127; blue, 229 }  ,draw opacity=1 ]   (191.48,60.94) -- (209.56,109.06) ;
\draw [color={rgb, 255:red, 40; green, 127; blue, 229 }  ,draw opacity=1 ]   (221.48,60.94) -- (239.56,109.06) ;
\draw [color={rgb, 255:red, 40; green, 127; blue, 229 }  ,draw opacity=1 ]   (209.56,60.94) -- (191.48,109.2) ;
\draw [color={rgb, 255:red, 40; green, 127; blue, 229 }  ,draw opacity=1 ]   (239.52,60.94) -- (221.44,109.2) ;
\draw [color={rgb, 255:red, 40; green, 127; blue, 229 }  ,draw opacity=1 ]   (191.48,90.94) -- (239.56,109.06) ;
\draw [color={rgb, 255:red, 40; green, 127; blue, 229 }  ,draw opacity=1 ]   (239.52,60.94) -- (191.48,79.06) ;
\draw [color={rgb, 255:red, 40; green, 127; blue, 229 }  ,draw opacity=1 ]   (239.56,90.94) -- (191.48,109.2) ;
\draw [color={rgb, 255:red, 40; green, 127; blue, 229 }  ,draw opacity=1 ]   (185.54,47.13) .. controls (199.88,30.5) and (230.38,30) .. (245.5,47.13) ;
\draw [color={rgb, 255:red, 40; green, 127; blue, 229 }  ,draw opacity=1 ]   (185.54,77.13) .. controls (199.88,60.5) and (230.38,60) .. (245.5,77.13) ;
\draw [color={rgb, 255:red, 40; green, 127; blue, 229 }  ,draw opacity=1 ]   (253.37,55) .. controls (270.38,70) and (270.38,99.5) .. (253.37,115.14) ;
\draw [color={rgb, 255:red, 40; green, 127; blue, 229 }  ,draw opacity=1 ]   (177.66,55) .. controls (162.88,68) and (161.38,99.5) .. (177.66,115.14) ;
\draw [color={rgb, 255:red, 40; green, 127; blue, 229 }  ,draw opacity=1 ]   (185.54,123.02) .. controls (199.88,139) and (229.38,140.5) .. (245.5,123.02) ;
\draw [color={rgb, 255:red, 40; green, 127; blue, 229 }  ,draw opacity=1 ]   (223.38,55) .. controls (240.38,70) and (240.38,99.5) .. (223.38,115.14) ;
\draw  [color={rgb, 255:red, 203; green, 0; blue, 23 }  ,draw opacity=1 ] (187.52,57.05) -- (243.42,57.05) -- (243.42,112.95) -- (187.52,112.95) -- cycle ;
\end{tikzpicture}
\caption{Clique pooling for a regular graph. The left graph represents a 4 pixel by 4 pixel image with nodes and edges in black and blue squares showing the maximal cliques. The right graph is the result of pooling where the maximal cliques have become nodes. Note that the neighbourhoods of each node have grown to include their neighbours' neighbours, resulting in a maximal clique of nine nodes shown by the red square.}
\label{regular_grid_pooling}
\end{center}
\vskip -0.5in
\end{wrapfigure}
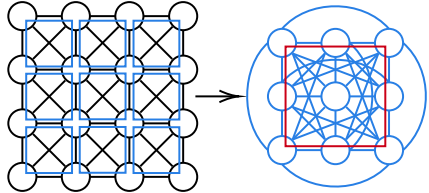

Images can be thought of as highly regular graphs. Indeed, this is the primary motivation for most work in the field of graph neural networks \citep{Kipf2016Semi-SupervisedNetworks,Defferrard2016ConvolutionalFiltering} and often image recognition techniques are straightforwardly adapted to graphs with great success \citep{Velickovic2017GraphNetworks,Cangea2018TowardsClassifiers}. Our method goes against the grain in this sense as, whilst it is based on concepts native to irregular graphs, we are able to apply it to images without issue.

To do this we must first define the graph structure for images. We argue that the use of 3-by-3 convolutions implies the graph structure as shown in figure \ref{regular_grid_pooling}, that is, pixels are connected to their eight immediate neighbours. As the figure shows, the first pool will be 2-by-2 with a stride of 1. The second pool will then be 3-by-3 with a stride of 1. 
We can analyse how this progresses by considering the 1-dimensional case, without loss of generality, as the 2D case is no more complicated than the same thing happening in two directions at the once\footnote{And the same can be said for an N-dimensional grid.}.

\begin{remark}[Length Reduction]
Applying clique-pool $n$-times on a 1-dimensional grid (a chain) reduces the length of the grid by $r_n=2^n - 1$. \footnote{See Appendix \ref{appenC} for proof and illustration.}
\end{remark}

In a typical \textsc{cnn}, the architecture will be such that the pools reduce the input to a single pixel in the spatial dimension by the final layer, using $(2\times2)$ pooling to do so. The length can be expressed in terms of the number of pools, $n$, as $L = 2^n$ and therefore, conveniently,
\begin{align*}
    L - r_n = L - 2^n + 1 = 1
\end{align*}
so the same number of clique-pooling operations will reduce the input to a single spatial dimension. As such the clique pool operation can be substituted into existing \textsc{cnn} architectures directly, replacing pools without any additional changes.


\section{Experimental Setup \& Results}
\label{expSetup}
We test the approach extensively on irregular graph classification problems and also demonstrate how the pooling works being substituted into a standard \textsc{cnn} architecture in the \textsc{vgg}-form.

\begin{table*}[ht!]
\label{irreg_results}
\vskip 0.15in
\begin{center}
\begin{small}
\begin{sc}
\begin{tabular}{lcccr}
\toprule
 & \multicolumn{4}{c}{Datasets} \\
 \cmidrule{2-5}
Model & Enzymes & DD & Proteins & Collab \\
\midrule
Graphlet & 41.03 & 74.85 & 64.66 & 72.91 \\ 
Shortest-path & 42.32 & 78.86 & 59.10 & \textbf{76.43} \\
1-WL & 53.43 & 74.02 & 78.61 & 73.76 \\
WL-QA & 60.13 & 79.04 & 80.74 & 75.26 \\
\midrule
PatchySAN & - & 76.27 & 72.60 & 75.00 \\
GraphSAGE & 54.25 & 75.42 & 68.25 & 70.48 \\
ECC & 53.50 & 74.10 & 67.79 & 72.65 \\
Set2Set & 60.15 & 78.12 & 71.75 & 74.29 \\
SortPool & 57.12 & 79.37 & 73.76 & 75.54 \\
DiffPool-Det & 58.33 & 75.47 & \textbf{82.13} & 75.62 \\
DiffPool-NoLP & 61.95 & 79.98 & 76.22 & 75.58 \\
DiffPool & 62.53 & \textbf{80.64} & 76.25 & 75.48 \\
Sparse HGC \cite{Cangea2018TowardsClassifiers} & \textbf{64.17} & 78.59 & 74.54 & 75.46 \\
\midrule
 CliquePool (ours) & 60.71 & 77.33 & 72.59 & 74.50 \\
\bottomrule
\end{tabular}
\end{sc}
\end{small}
\end{center}
\caption{Classification accuracy percentages.}

\end{table*}

 We benchmark the performance on irregular graphs on standard datasets. For \textit{Enzymes, DD, Proteins} we use the same architecture: two blocks of Graph Convolution Network + Pool followed by a convolutional layer. We use the mean as the readout function. For \textit{Enzymes} we use Graph\textsc{sage} (`mean' variant) for the convolutions, and \textsc{gcn} for others. The outputs of each layer are normalised by the $L^2$-norm. For \textit{Collab} we use the \textsc{gcn} and one layer of pooling, followed by another convolution layer. To provide a comparison of the number of pooling parameters, in the case of \textit{DD}, DiffPool requires $27,776$ parameters, Graph U-Net requires $192$ whilst our method is nonparametric (0 parameters).
We use $128$ hidden units for \textit{Enzymes}, and $64$ for the other datasets. We use the Adam optimizer (learning rate of $0.0001$, weight decay of $0.001$). We train for $1000$ epochs for \textit{Enzymes}, $100$ for \textit{DD}, $500$ for \textit{Proteins}, and $200$ for \textit{Collab}. 


Regular graph pooling is tested against the \textsc{cifar}-10 benchmark using the standard architecture proposed by \citealt{Jetley2018LearnAttention}, an adaptation of the \textsc{vgg} ImageNet entry \cite{Simonyan2015VeryRecognition}. The architecture consists of alternating stacks of convolutional layers, zero-padded to maintain spatial dimensions, and $(2\times2)$ max-pools with a stride of $2$ to reduce the spatial dimensions by half, with a final multilayer perceptron (\textsc{mlp}) classifier\footnote{Full specification is given in appendix \ref{net_arch}}.
C\textsc{ifar} images are $(32,32,3)$ so five stride-2 pools are needed to reduce to $(1,1,F)$. Following the derivation in section \ref{applicationImages}, these are replaced by $(2\times2),(3\times3),(5\times5),(9\times9)$ \& $(17\times17)$ pools all with stride of 1. 

\begin{wraptable}{r}{5.5cm}
\begin{center}
\begin{small}
\begin{sc}
\begin{tabular}{lrr}
\toprule
\multicolumn{1}{l}{Model} & Mean          & Std. Dev. \\
\midrule
2-by-2 pool               & \textbf{92.4} & 0.3       \\
Clique pool               & 92.0          & 0.5       \\
\bottomrule
\end{tabular}
\caption{\textsc{cifar}-10 results}
\end{sc}
\end{small}
\end{center}
\label{cifar10res}
\vskip 0.3in
\end{wraptable}


Our experiments show that our non-parametric approach is competitive with the parametric approaches, as detailed in table \ref{irreg_results}. The method outperforms the Graph\textsc{sage} baseline and most of the kernel-based and \textsc{gnn} approaches. Moreover, because we do not introduce any additional parameters our method is fast to train and does not suffer the instabilities associated with \textsc{DiffPool}. It also outperforms the \textsc{DiffPool} with deterministic clustering on two datasets. The image investigation found a small, but significant ($p = 0.02$), reduction of in mean accuracy over the 2-\textsc{by}-2 pool baseline in a 10-fold cross-validation comparison, presented in table \ref{cifar10res}.

\addtocounter{remark}{-2}

\bibliography{iclr2019_conference,mendeley2}

\begin{thebibliography}{20}
\providecommand{\natexlab}[1]{#1}
\providecommand{\url}[1]{\texttt{#1}}
\expandafter\ifx\csname urlstyle\endcsname\relax
  \providecommand{\doi}[1]{doi: #1}\else
  \providecommand{\doi}{doi: \begingroup \urlstyle{rm}\Url}\fi

\bibitem[Bron \& Kerbosch(1973)Bron and Kerbosch]{Bron1973AlgorithmGraph}
Coen Bron and Joep Kerbosch.
\newblock {Algorithm 457: finding all cliques of an undirected graph}.
\newblock \emph{Communications of the ACM}, 16\penalty0 (9):\penalty0 575--577,
  9 1973.
\newblock ISSN 00010782.
\newblock \doi{10.1145/362342.362367}.
\newblock URL \url{http://portal.acm.org/citation.cfm?doid=362342.362367}.

\bibitem[Bronstein et~al.(2017)Bronstein, LeCun, Szlam, Vandergheynst, and
  Bruna]{Bronstein2017GeometricData}
Michael~M. Bronstein, Yann LeCun, Arthur Szlam, Pierre Vandergheynst, and Joan
  Bruna.
\newblock {Geometric Deep Learning: Going beyond Euclidean data}.
\newblock \emph{IEEE Signal Processing Magazine}, 34\penalty0 (4):\penalty0
  18--42, 11 2017.
\newblock ISSN 1053-5888.
\newblock \doi{10.1109/msp.2017.2693418}.
\newblock URL \url{http://arxiv.org/abs/1611.08097
  http://dx.doi.org/10.1109/MSP.2017.2693418}.

\bibitem[Cangea et~al.(2018)Cangea, Veli{\v{c}}kovi{\'{c}}, Jovanovi{\'{c}},
  Kipf, and Li{\`{o}}]{Cangea2018TowardsClassifiers}
Cătălina Cangea, Petar Veli{\v{c}}kovi{\'{c}}, Nikola Jovanovi{\'{c}}, Thomas
  Kipf, and Pietro Li{\`{o}}.
\newblock {Towards Sparse Hierarchical Graph Classifiers}.
\newblock 11 2018.
\newblock URL \url{http://arxiv.org/abs/1811.01287}.

\bibitem[Conte et~al.()Conte, De~Virgilio, Maccioni, Patrignani, and
  Torlone]{ConteFindingNetworks}
Alessio Conte, Roberto De~Virgilio, Antonio Maccioni, Maurizio Patrignani, and
  Riccardo Torlone.
\newblock {Finding All Maximal Cliques in Very Large Social Networks}.
\newblock \doi{10.5441/002/edbt.2016.18}.
\newblock URL \url{https://openproceedings.org/2016/conf/edbt/paper-54.pdf}.

\bibitem[Defferrard et~al.(2016)Defferrard, Bresson, and
  Vandergheynst]{Defferrard2016ConvolutionalFiltering}
Michaël Defferrard, Xavier Bresson, and Pierre Vandergheynst.
\newblock {Convolutional Neural Networks on Graphs with Fast Localized Spectral
  Filtering}.
\newblock 6 2016.
\newblock URL \url{http://arxiv.org/abs/1606.09375}.

\bibitem[Duvenaud et~al.(2015)Duvenaud, Maclaurin, Aguilera-Iparraguirre,
  G{\'{o}}mez-Bombarelli, Hirzel, Aspuru-Guzik, and
  Adams]{Duvenaud2015ConvolutionalFingerprints}
David Duvenaud, Dougal Maclaurin, Jorge Aguilera-Iparraguirre, Rafael
  G{\'{o}}mez-Bombarelli, Timothy Hirzel, Alán Aspuru-Guzik, and Ryan~P.
  Adams.
\newblock {Convolutional Networks on Graphs for Learning Molecular
  Fingerprints}.
\newblock 9 2015.
\newblock URL \url{http://arxiv.org/abs/1509.09292}.

\bibitem[Eppstein \& Strash(2011)Eppstein and
  Strash]{Eppstein2011ListingGraphs}
David Eppstein and Darren Strash.
\newblock {Listing All Maximal Cliques in Large Sparse Real-World Graphs}.
\newblock 3 2011.
\newblock URL \url{http://arxiv.org/abs/1103.0318}.

\bibitem[Eppstein et~al.(2010)Eppstein, L{\"{o}}ffler, and
  Strash]{Eppstein2010ListingTime}
David Eppstein, Maarten L{\"{o}}ffler, and Darren Strash.
\newblock {Listing All Maximal Cliques in Sparse Graphs in Near-optimal Time}.
\newblock 6 2010.
\newblock URL \url{http://arxiv.org/abs/1006.5440}.

\bibitem[Gao \& Ji(2018)Gao and Ji]{gao2018graph}
Hongyang Gao and Shuiwang Ji.
\newblock Graph u-net.
\newblock 2018.

\bibitem[Gilmer et~al.(2017)Gilmer, Schoenholz, Riley, Vinyals, and
  Dahl]{Gilmer2017NeuralChemistry}
Justin Gilmer, Samuel~S. Schoenholz, Patrick~F. Riley, Oriol Vinyals, and
  George~E. Dahl.
\newblock {Neural Message Passing for Quantum Chemistry}.
\newblock 4 2017.
\newblock URL \url{http://arxiv.org/abs/1704.01212}.

\bibitem[Ioffe \& Szegedy(2015)Ioffe and Szegedy]{Ioffe2015BatchShift}
Sergey Ioffe and Christian Szegedy.
\newblock {Batch Normalization: Accelerating Deep Network Training by Reducing
  Internal Covariate Shift}.
\newblock 2 2015.
\newblock URL \url{http://arxiv.org/abs/1502.03167}.

\bibitem[Jetley et~al.(2018)Jetley, Lord, Lee, and
  Torr]{Jetley2018LearnAttention}
Saumya Jetley, Nicholas~A. Lord, Namhoon Lee, and Philip H.~S. Torr.
\newblock {Learn To Pay Attention}.
\newblock 4 2018.
\newblock URL \url{https://arxiv.org/abs/1804.02391}.

\bibitem[Kipf et~al.(2018)Kipf, Fetaya, Wang, Welling, and
  Zemel]{Kipf2018NeuralSystems}
Thomas Kipf, Ethan Fetaya, Kuan-Chieh Wang, Max Welling, and Richard Zemel.
\newblock {Neural Relational Inference for Interacting Systems}.
\newblock 2 2018.
\newblock URL \url{http://arxiv.org/abs/1802.04687}.

\bibitem[Kipf \& Welling(2016)Kipf and
  Welling]{Kipf2016Semi-SupervisedNetworks}
Thomas~N. Kipf and Max Welling.
\newblock {Semi-Supervised Classification with Graph Convolutional Networks}.
\newblock 9 2016.
\newblock URL \url{http://arxiv.org/abs/1609.02907}.

\bibitem[Moon \& Moser(1965)Moon and Moser]{Moon1965OnGraphs}
J.~W. Moon and L.~Moser.
\newblock {On cliques in graphs}.
\newblock \emph{Israel Journal of Mathematics}, 3\penalty0 (1):\penalty0
  23--28, 3 1965.
\newblock ISSN 0021-2172.
\newblock \doi{10.1007/BF02760024}.
\newblock URL \url{http://link.springer.com/10.1007/BF02760024}.

\bibitem[Simonyan \& Zisserman(2015)Simonyan and
  Zisserman]{Simonyan2015VeryRecognition}
Karen Simonyan and Andrew Zisserman.
\newblock {Very Deep Convolutional Networks for Large-Scale Image Recognition}.
\newblock Technical report, 2015.
\newblock URL \url{http://www.robots.ox.ac.uk/}.

\bibitem[Tomita et~al.(2006)Tomita, Tanaka, and Takahashi]{tomita2006worst}
Etsuji Tomita, Akira Tanaka, and Haruhisa Takahashi.
\newblock The worst-case time complexity for generating all maximal cliques and
  computational experiments.
\newblock \emph{Theoretical Computer Science}, 363\penalty0 (1):\penalty0
  28--42, 2006.

\bibitem[Veli{\v{c}}kovi{\'{c}} et~al.(2017)Veli{\v{c}}kovi{\'{c}}, Cucurull,
  Casanova, Romero, Li{\`{o}}, and Bengio]{Velickovic2017GraphNetworks}
Petar Veli{\v{c}}kovi{\'{c}}, Guillem Cucurull, Arantxa Casanova, Adriana
  Romero, Pietro Li{\`{o}}, and Yoshua Bengio.
\newblock {Graph Attention Networks}.
\newblock 10 2017.
\newblock URL \url{http://arxiv.org/abs/1710.10903}.

\bibitem[Ying et~al.(2018)Ying, You, Morris, Ren, Hamilton, and
  Leskovec]{Ying2018HierarchicalPooling}
Rex Ying, Jiaxuan You, Christopher Morris, Xiang Ren, William~L Hamilton, and
  Jure Leskovec.
\newblock {Hierarchical Graph Representation Learning with Differentiable
  Pooling}.
\newblock 2018.
\newblock ISSN 18160948.
\newblock \doi{arXiv:1806.08804v3}.
\newblock URL \url{https://arxiv.org/pdf/1806.08804.pdf
  http://arxiv.org/abs/1806.08804}.

\bibitem[Zhang et~al.(2005)Zhang, Abu-Khzam, Baldwin, Chesler, Langston, and
  Samatova]{Zhang2005Genome-scaleBiology}
Yun Zhang, Faisal~N. Abu-Khzam, Nicole~E. Baldwin, Elissa~J. Chesler,
  Michael~A. Langston, and Nagiza~F. Samatova.
\newblock {Genome-scale computational approaches to memory-intensive
  applications in systems biology}.
\newblock In \emph{Proceedings of the ACM/IEEE 2005 Supercomputing Conference,
  SC'05}, 2005.
\newblock ISBN 1595930612.
\newblock \doi{10.1109/SC.2005.29}.

\end{thebibliography}
\bibliographystyle{iclr2019_conference}

\clearpage
\appendix
\section{Network Architectures}
\label{net_arch}
In the interests of clarity and reproducibility we present here the network architectures used in our investigations. We believe the following to be a complete description though the authors would gladly welcome any correspondence requesting further specification, clarification or suggestions relating to improving these descriptions.

For compactness, we've used abbreviations. \textsc{Conv.*} is a block consisting of a convolutional layer with a $3\times3$ kernel, batch-normalization \citep{Ioffe2015BatchShift} with numerical stabilisation ($\epsilon = 1\times10^-5$) and momentum of $0.1$ for the affine transform parameters ($\gamma,\beta$), with a \textsc{r}e\textsc{lu} activation. `$\times$ N' refers to how many times this block is repeated, which could also be inferred by from the numbering in the \textsc{Layer} column. \textsc{Flatten} returns the layer input with dimensions of size 1 removed, sometimes called \textit{squeezing}. \textsc{Linear} layers are fully-connected and followed by a \textsc{r}e\textsc{lu} activation except in the final layer where a softmax is used. For pooling layers the size is given as $(\Delta x \times \Delta y), stride$.

\begin{table}[h]
\begin{center}
\begin{small}
\begin{sc}
\begin{tabular}{llcr}
\toprule
\textbf{Layer} & \textbf{Type}     & \textbf{Size/Features} & \textbf{Shape} \\
\midrule
Input          & -                 & -             & (32,32,3)      \\
1,2            & conv.* $\times$ 2 & 64            & (32,32,64)     \\
-              & maxpool              & $(2\times2)$, 2    & (16,16,64)     \\
3,4            & conv.* $\times$ 2 & 128           & (16,16,128)    \\
-              & maxpool              & $(2\times2)$, 2     & (8,8,128)      \\
5,6,7          & conv.* $\times$ 3 & 256           & (8,8,256)      \\
-              & maxpool              & $(2\times2)$, 2     & (4,4,256)      \\
8,9,10         & conv.* $\times$ 3 & 512           & (4,4,512)      \\
-              & maxpool              & $(2\times2)$, 2     & (2,2,512)      \\
11,12,13       & conv.* $\times$ 3 & 512           & (2,2,512)      \\
-              & maxpool              & $(2\times2)$, 2     & (1,1,512)      \\
-              & flatten           & -             & (512)          \\
14             & linear            & 512           & (512)          \\
-              & dropout           & $p=0.3$       & (512)          \\
15             & linear            & 10            & (10)           \\
\bottomrule
\end{tabular}
\caption{Baseline VGG-like architecture.}
\quad
\end{sc}
\end{small}
\end{center}
\begin{center}
\begin{small}
\begin{sc}
\begin{tabular}{llcr}
\toprule
\textbf{Layer} & \textbf{Type}     & \textbf{Size/Features} & \textbf{Shape} \\
\midrule
Input          & -                 & -                 & (32,32,3)      \\
1,2            & conv.* $\times$ 2 & 64                & (32,32,64)     \\
-              & maxpool           & $(2\times2)$, 1   & (31,31,64)     \\
3,4            & conv.* $\times$ 2 & 128               & (31,31,128)    \\
-              & maxpool           & $(3\times3)$, 1   & (29,29,128)    \\
5,6,7          & conv.* $\times$ 3 & 256               & (29,29,256)    \\
-              & maxpool           & $(5\times5)$, 1   & (25,25,256)    \\
8,9,10         & conv.* $\times$ 3 & 512               & (25,25,512)    \\
-              & maxpool           & $(9\times9)$, 1   & (17,17,512)    \\
11,12,13       & conv.* $\times$ 3 & 512               & (17,17,512)    \\
-              & maxpool           & $(17\times17)$, 1 & (1,1,512)      \\
-              & flatten           & -                 & (512)          \\
14             & linear            & 512               & (512)          \\
-              & dropout           & $p=0.3$           & (512)          \\
15             & linear            & 10                & (10)           \\ 
\bottomrule
\end{tabular}
\caption{Clique-pooled VGG-like architecture.}

\end{sc}
\end{small}
\end{center}
\end{table}

\section{Maximal Cliques}
The enumeration of maximal cliques has been a core component in gene expression networks
analysis, cis regulatory motif finding, and the study of quantitative trait loci for high-throughput molecular
phenotypes \cite{Zhang2005Genome-scaleBiology}. As such, significant effort has gone to developing efficient methods of enumerating all the maximal cliques in a graph. The upper bound of the number of cliques is $3^{n/3}$. The algorithm for finding maximal cliques presented by \citealt{Bron1973AlgorithmGraph}, was adapted by \citealt{tomita2006worst} to find the maximal cliques in an iterative way without having to store previous cliques or many candidates in memory. \cite{Zhang2005Genome-scaleBiology} showed a way of enumerating the cliques, sorted by size, in a parallelized way. While their primary focus is parallelizing this method in shared-memory machines, they show that it is possible to do the same in distributed machines. They also demonstrate that it is possible to load-balance the sub-tasks efficiently, showing a near ideal relative speedup (defined as the ratio between $2p$ processors and $p$ processors run times).  
Finally, several modifications have been made showing improved performance on large real-world graphs (including social graphs) \cite{Eppstein2010ListingTime,Eppstein2011ListingGraphs,ConteFindingNetworks}

\section{Proofs}
\label{appenC}
\subsection{Convergence}

We recall,

\begin{remark}[Convergence]
Given a connected and finite graph, the clique-pooling operator converges to a single node after finitely many steps.
\end{remark}

\begin{proof}
Consider the shortest-path between two nodes, $d(u,v)$, and the corresponding distance in the new graph, $d(u',v')$ where $u'$ and $v'$ are the most distant cliques containing $u$ and $v$, respectively. For each assigned maximal clique that the path between $u$ and $v$ traverses, $d(u',v')$ is reduced from $d(u,v)$ by 1. If the path does not traverse any pooled cliques then the distance remains constant as the path will be unchanged. As there is always a largest clique, the distance between some nodes is always reduced in the newly formed graph. Therefore the sum of the distances between all pairs of original nodes \textit{must} decrease in each pooling. As the graph is finite and connected the sum of the distances between nodes must also be finite and so will reduce to 0, a single node, within a finite number of pooling operations. 
\end{proof}
However, it is possible to construct graphs that will initially grow in the number of nodes through pooling, the most straightforward example being a bipartite graph. In this case every pair of nodes sharing an edge forms a maximal clique, all of size 2 and thus all equally large. The pooled-graph then has as many nodes as the original graph had edges. In the worst possible case, the number of nodes in the pooled graph will balloon to the maximal clique limit, $\mathcal{O}(3^{\sfrac{n}{3}})$ \citep{Moon1965OnGraphs}, although this will be fully-connected and collapse immediately.

\subsection{Receptive field}
We recall, 
\begin{remark}[Length Reduction]
Applying clique-pool $n$-times on a 1-dimensional grid (a chain) reduces the length of the grid by $r_n=2^n - 1$.
\end{remark}

\begin{figure}[h]
\begin{center}

\tikzset{every picture/.style={line width=0.75pt}} 

\begin{tikzpicture}[x=0.75pt,y=0.75pt,yscale=-1,xscale=1]

\draw [color={rgb, 255:red, 40; green, 127; blue, 229 }  ,draw opacity=1 ]   (94.5,98) -- (136.5,105) ;

\draw [color={rgb, 255:red, 40; green, 127; blue, 229 }  ,draw opacity=1 ]   (108.5,98) -- (136.5,105) ;

\draw [color={rgb, 255:red, 40; green, 127; blue, 229 }  ,draw opacity=1 ]   (123,98) -- (137,105) ;

\draw [color={rgb, 255:red, 40; green, 127; blue, 229 }  ,draw opacity=1 ]   (136.5,98) -- (136.5,105) ;

\draw [color={rgb, 255:red, 40; green, 127; blue, 229 }  ,draw opacity=1 ]   (136.5,105) -- (150.5,98) ;

\draw [color={rgb, 255:red, 40; green, 127; blue, 229 }  ,draw opacity=1 ]   (136.5,105) -- (164.5,98) ;

\draw [color={rgb, 255:red, 40; green, 127; blue, 229 }  ,draw opacity=1 ] [dash pattern={on 0.84pt off 2.51pt}]  (137,105) -- (179,98) ;

\draw [color={rgb, 255:red, 40; green, 127; blue, 229 }  ,draw opacity=1 ]   (199.5,91) -- (199.5,84) ;

\draw [color={rgb, 255:red, 40; green, 127; blue, 229 }  ,draw opacity=1 ]   (213.5,77) -- (199.5,70) ;

\draw [color={rgb, 255:red, 40; green, 127; blue, 229 }  ,draw opacity=1 ]   (199.5,77) -- (199.5,70) ;

\draw [color={rgb, 255:red, 40; green, 127; blue, 229 }  ,draw opacity=1 ]   (199.5,63) -- (206.5,56) ;

\draw [color={rgb, 255:red, 40; green, 127; blue, 229 }  ,draw opacity=1 ]   (199.5,77) -- (213.5,70) ;

\draw [color={rgb, 255:red, 40; green, 127; blue, 229 }  ,draw opacity=1 ]   (199.5,91) -- (213.5,84) ;

\draw [color={rgb, 255:red, 40; green, 127; blue, 229 }  ,draw opacity=1 ]   (199.5,91) -- (227.5,84) ;

\draw [color={rgb, 255:red, 40; green, 127; blue, 229 }  ,draw opacity=1 ]   (38.5,70) -- (52.5,77) ;

\draw [color={rgb, 255:red, 40; green, 127; blue, 229 }  ,draw opacity=1 ]   (52.5,70) -- (66.5,77) ;

\draw [color={rgb, 255:red, 40; green, 127; blue, 229 }  ,draw opacity=1 ]   (66.5,70) -- (80.5,77) ;

\draw [color={rgb, 255:red, 40; green, 127; blue, 229 }  ,draw opacity=1 ]   (80.5,70) -- (94.5,77) ;

\draw [color={rgb, 255:red, 40; green, 127; blue, 229 }  ,draw opacity=1 ]   (94.5,70) -- (108.5,77) ;

\draw [color={rgb, 255:red, 40; green, 127; blue, 229 }  ,draw opacity=1 ]   (108.5,70) -- (122.5,77) ;

\draw [color={rgb, 255:red, 40; green, 127; blue, 229 }  ,draw opacity=1 ]   (122.5,70) -- (136.5,77) ;

\draw [color={rgb, 255:red, 40; green, 127; blue, 229 }  ,draw opacity=1 ]   (136.5,70) -- (150.5,77) ;

\draw [color={rgb, 255:red, 40; green, 127; blue, 229 }  ,draw opacity=1 ]   (150.5,70) -- (164.5,77) ;

\draw [color={rgb, 255:red, 40; green, 127; blue, 229 }  ,draw opacity=1 ]   (213.5,70) -- (227.5,77) ;

\draw [color={rgb, 255:red, 40; green, 127; blue, 229 }  ,draw opacity=1 ]   (241.5,70) -- (227.5,77) ;

\draw [color={rgb, 255:red, 40; green, 127; blue, 229 }  ,draw opacity=1 ]   (227.5,70) -- (213.5,77) ;

\draw [color={rgb, 255:red, 40; green, 127; blue, 229 }  ,draw opacity=1 ]   (164.5,70) -- (150.5,77) ;

\draw [color={rgb, 255:red, 40; green, 127; blue, 229 }  ,draw opacity=1 ]   (150.5,70) -- (136.5,77) ;

\draw [color={rgb, 255:red, 40; green, 127; blue, 229 }  ,draw opacity=1 ]   (136.5,70) -- (122.5,77) ;

\draw [color={rgb, 255:red, 40; green, 127; blue, 229 }  ,draw opacity=1 ]   (122.5,70) -- (108.5,77) ;

\draw [color={rgb, 255:red, 40; green, 127; blue, 229 }  ,draw opacity=1 ]   (108.5,70) -- (94.5,77) ;

\draw [color={rgb, 255:red, 40; green, 127; blue, 229 }  ,draw opacity=1 ]   (94.5,70) -- (80.5,77) ;

\draw [color={rgb, 255:red, 40; green, 127; blue, 229 }  ,draw opacity=1 ]   (80.5,70) -- (66.5,77) ;

\draw [color={rgb, 255:red, 40; green, 127; blue, 229 }  ,draw opacity=1 ]   (66.5,70) -- (52.5,77) ;

\draw [color={rgb, 255:red, 40; green, 127; blue, 229 }  ,draw opacity=1 ]   (52.5,70) -- (52.5,77) ;

\draw [color={rgb, 255:red, 40; green, 127; blue, 229 }  ,draw opacity=1 ]   (66.5,70) -- (66.5,77) ;

\draw [color={rgb, 255:red, 40; green, 127; blue, 229 }  ,draw opacity=1 ]   (80.5,70) -- (80.5,77) ;

\draw [color={rgb, 255:red, 40; green, 127; blue, 229 }  ,draw opacity=1 ]   (94.5,70) -- (94.5,77) ;

\draw [color={rgb, 255:red, 40; green, 127; blue, 229 }  ,draw opacity=1 ]   (108.5,70) -- (108.5,77) ;

\draw [color={rgb, 255:red, 40; green, 127; blue, 229 }  ,draw opacity=1 ]   (122.5,70) -- (122.5,77) ;

\draw [color={rgb, 255:red, 40; green, 127; blue, 229 }  ,draw opacity=1 ]   (136.5,70) -- (136.5,77) ;

\draw [color={rgb, 255:red, 40; green, 127; blue, 229 }  ,draw opacity=1 ]   (150.5,70) -- (150.5,77) ;

\draw [color={rgb, 255:red, 40; green, 127; blue, 229 }  ,draw opacity=1 ]   (164.5,70) -- (164.5,77) ;

\draw [color={rgb, 255:red, 40; green, 127; blue, 229 }  ,draw opacity=1 ]   (213.5,70) -- (213.5,77) ;

\draw [color={rgb, 255:red, 40; green, 127; blue, 229 }  ,draw opacity=1 ]   (227.5,70) -- (227.5,77) ;

\draw [color={rgb, 255:red, 40; green, 127; blue, 229 }  ,draw opacity=1 ]   (31.5,56) -- (38.5,63) ;

\draw [color={rgb, 255:red, 40; green, 127; blue, 229 }  ,draw opacity=1 ]   (45.5,56) -- (38.5,63) ;

\draw [color={rgb, 255:red, 40; green, 127; blue, 229 }  ,draw opacity=1 ]   (45.5,56) -- (52.5,63) ;

\draw [color={rgb, 255:red, 40; green, 127; blue, 229 }  ,draw opacity=1 ]   (59.5,56) -- (52.5,63) ;

\draw [color={rgb, 255:red, 40; green, 127; blue, 229 }  ,draw opacity=1 ]   (59.5,56) -- (66.5,63) ;

\draw [color={rgb, 255:red, 40; green, 127; blue, 229 }  ,draw opacity=1 ]   (73.5,56) -- (80.5,63) ;

\draw [color={rgb, 255:red, 40; green, 127; blue, 229 }  ,draw opacity=1 ]   (87.5,56) -- (94.5,63) ;

\draw [color={rgb, 255:red, 40; green, 127; blue, 229 }  ,draw opacity=1 ]   (101.5,56) -- (108.5,63) ;

\draw [color={rgb, 255:red, 40; green, 127; blue, 229 }  ,draw opacity=1 ]   (115.5,56) -- (122.5,63) ;

\draw [color={rgb, 255:red, 40; green, 127; blue, 229 }  ,draw opacity=1 ]   (129.5,56) -- (136.5,63) ;

\draw [color={rgb, 255:red, 40; green, 127; blue, 229 }  ,draw opacity=1 ]   (143.5,56) -- (150.5,63) ;

\draw [color={rgb, 255:red, 40; green, 127; blue, 229 }  ,draw opacity=1 ]   (157.5,56) -- (164.5,63) ;

\draw [color={rgb, 255:red, 40; green, 127; blue, 229 }  ,draw opacity=1 ]   (206.5,56) -- (213.5,63) ;

\draw [color={rgb, 255:red, 40; green, 127; blue, 229 }  ,draw opacity=1 ]   (220.5,56) -- (227.5,63) ;

\draw [color={rgb, 255:red, 40; green, 127; blue, 229 }  ,draw opacity=1 ]   (234.5,56) -- (241.5,63) ;

\draw [color={rgb, 255:red, 40; green, 127; blue, 229 }  ,draw opacity=1 ]   (73.5,56) -- (66.5,63) ;

\draw [color={rgb, 255:red, 40; green, 127; blue, 229 }  ,draw opacity=1 ]   (87.5,56) -- (80.5,63) ;

\draw [color={rgb, 255:red, 40; green, 127; blue, 229 }  ,draw opacity=1 ]   (101.5,56) -- (94.5,63) ;

\draw [color={rgb, 255:red, 40; green, 127; blue, 229 }  ,draw opacity=1 ]   (115.5,56) -- (108.5,63) ;

\draw [color={rgb, 255:red, 40; green, 127; blue, 229 }  ,draw opacity=1 ]   (129.5,56) -- (122.5,63) ;

\draw [color={rgb, 255:red, 40; green, 127; blue, 229 }  ,draw opacity=1 ]   (143.5,56) -- (136.5,63) ;

\draw [color={rgb, 255:red, 40; green, 127; blue, 229 }  ,draw opacity=1 ]   (157.5,56) -- (150.5,63) ;

\draw [color={rgb, 255:red, 40; green, 127; blue, 229 }  ,draw opacity=1 ]   (171.5,56) -- (164.5,63) ;

\draw [color={rgb, 255:red, 40; green, 127; blue, 229 }  ,draw opacity=1 ]   (220.5,56) -- (213.5,63) ;

\draw [color={rgb, 255:red, 40; green, 127; blue, 229 }  ,draw opacity=1 ]   (234.5,56) -- (227.5,63) ;

\draw [color={rgb, 255:red, 40; green, 127; blue, 229 }  ,draw opacity=1 ]   (248.5,56) -- (241.5,63) ;

\draw  [color={rgb, 255:red, 0; green, 0; blue, 0 }  ,draw opacity=1 ] (28,52.5) .. controls (28,50.57) and (29.57,49) .. (31.5,49) .. controls (33.43,49) and (35,50.57) .. (35,52.5) .. controls (35,54.43) and (33.43,56) .. (31.5,56) .. controls (29.57,56) and (28,54.43) .. (28,52.5) -- cycle ;
\draw  [color={rgb, 255:red, 0; green, 0; blue, 0 }  ,draw opacity=1 ] (42,52.5) .. controls (42,50.57) and (43.57,49) .. (45.5,49) .. controls (47.43,49) and (49,50.57) .. (49,52.5) .. controls (49,54.43) and (47.43,56) .. (45.5,56) .. controls (43.57,56) and (42,54.43) .. (42,52.5) -- cycle ;
\draw  [color={rgb, 255:red, 0; green, 0; blue, 0 }  ,draw opacity=1 ] (56,52.5) .. controls (56,50.57) and (57.57,49) .. (59.5,49) .. controls (61.43,49) and (63,50.57) .. (63,52.5) .. controls (63,54.43) and (61.43,56) .. (59.5,56) .. controls (57.57,56) and (56,54.43) .. (56,52.5) -- cycle ;
\draw  [color={rgb, 255:red, 0; green, 0; blue, 0 }  ,draw opacity=1 ] (70,52.5) .. controls (70,50.57) and (71.57,49) .. (73.5,49) .. controls (75.43,49) and (77,50.57) .. (77,52.5) .. controls (77,54.43) and (75.43,56) .. (73.5,56) .. controls (71.57,56) and (70,54.43) .. (70,52.5) -- cycle ;
\draw  [color={rgb, 255:red, 0; green, 0; blue, 0 }  ,draw opacity=1 ] (84,52.5) .. controls (84,50.57) and (85.57,49) .. (87.5,49) .. controls (89.43,49) and (91,50.57) .. (91,52.5) .. controls (91,54.43) and (89.43,56) .. (87.5,56) .. controls (85.57,56) and (84,54.43) .. (84,52.5) -- cycle ;
\draw  [color={rgb, 255:red, 0; green, 0; blue, 0 }  ,draw opacity=1 ] (98,52.5) .. controls (98,50.57) and (99.57,49) .. (101.5,49) .. controls (103.43,49) and (105,50.57) .. (105,52.5) .. controls (105,54.43) and (103.43,56) .. (101.5,56) .. controls (99.57,56) and (98,54.43) .. (98,52.5) -- cycle ;
\draw  [color={rgb, 255:red, 0; green, 0; blue, 0 }  ,draw opacity=1 ] (112,52.5) .. controls (112,50.57) and (113.57,49) .. (115.5,49) .. controls (117.43,49) and (119,50.57) .. (119,52.5) .. controls (119,54.43) and (117.43,56) .. (115.5,56) .. controls (113.57,56) and (112,54.43) .. (112,52.5) -- cycle ;
\draw  [color={rgb, 255:red, 0; green, 0; blue, 0 }  ,draw opacity=1 ] (126,52.5) .. controls (126,50.57) and (127.57,49) .. (129.5,49) .. controls (131.43,49) and (133,50.57) .. (133,52.5) .. controls (133,54.43) and (131.43,56) .. (129.5,56) .. controls (127.57,56) and (126,54.43) .. (126,52.5) -- cycle ;
\draw  [color={rgb, 255:red, 0; green, 0; blue, 0 }  ,draw opacity=1 ] (140,52.5) .. controls (140,50.57) and (141.57,49) .. (143.5,49) .. controls (145.43,49) and (147,50.57) .. (147,52.5) .. controls (147,54.43) and (145.43,56) .. (143.5,56) .. controls (141.57,56) and (140,54.43) .. (140,52.5) -- cycle ;
\draw  [color={rgb, 255:red, 0; green, 0; blue, 0 }  ,draw opacity=1 ] (154,52.5) .. controls (154,50.57) and (155.57,49) .. (157.5,49) .. controls (159.43,49) and (161,50.57) .. (161,52.5) .. controls (161,54.43) and (159.43,56) .. (157.5,56) .. controls (155.57,56) and (154,54.43) .. (154,52.5) -- cycle ;
\draw  [color={rgb, 255:red, 0; green, 0; blue, 0 }  ,draw opacity=1 ] (168,52.5) .. controls (168,50.57) and (169.57,49) .. (171.5,49) .. controls (173.43,49) and (175,50.57) .. (175,52.5) .. controls (175,54.43) and (173.43,56) .. (171.5,56) .. controls (169.57,56) and (168,54.43) .. (168,52.5) -- cycle ;
\draw  [color={rgb, 255:red, 0; green, 0; blue, 0 }  ,draw opacity=1 ] (203,52.5) .. controls (203,50.57) and (204.57,49) .. (206.5,49) .. controls (208.43,49) and (210,50.57) .. (210,52.5) .. controls (210,54.43) and (208.43,56) .. (206.5,56) .. controls (204.57,56) and (203,54.43) .. (203,52.5) -- cycle ;
\draw  [color={rgb, 255:red, 0; green, 0; blue, 0 }  ,draw opacity=1 ] (217,52.5) .. controls (217,50.57) and (218.57,49) .. (220.5,49) .. controls (222.43,49) and (224,50.57) .. (224,52.5) .. controls (224,54.43) and (222.43,56) .. (220.5,56) .. controls (218.57,56) and (217,54.43) .. (217,52.5) -- cycle ;
\draw  [color={rgb, 255:red, 0; green, 0; blue, 0 }  ,draw opacity=1 ] (231,52.5) .. controls (231,50.57) and (232.57,49) .. (234.5,49) .. controls (236.43,49) and (238,50.57) .. (238,52.5) .. controls (238,54.43) and (236.43,56) .. (234.5,56) .. controls (232.57,56) and (231,54.43) .. (231,52.5) -- cycle ;
\draw  [color={rgb, 255:red, 0; green, 0; blue, 0 }  ,draw opacity=1 ] (245,52.5) .. controls (245,50.57) and (246.57,49) .. (248.5,49) .. controls (250.43,49) and (252,50.57) .. (252,52.5) .. controls (252,54.43) and (250.43,56) .. (248.5,56) .. controls (246.57,56) and (245,54.43) .. (245,52.5) -- cycle ;
\draw [color={rgb, 255:red, 0; green, 0; blue, 0 }  ,draw opacity=1 ] [dash pattern={on 0.84pt off 2.51pt}]  (175,52.5) -- (203,52.5) ;

\draw [color={rgb, 255:red, 0; green, 0; blue, 0 }  ,draw opacity=1 ]   (35,52.5) -- (42,52.5) ;

\draw [color={rgb, 255:red, 0; green, 0; blue, 0 }  ,draw opacity=1 ]   (49,52.5) -- (56,52.5) ;

\draw [color={rgb, 255:red, 0; green, 0; blue, 0 }  ,draw opacity=1 ]   (63,52.5) -- (70,52.5) ;

\draw [color={rgb, 255:red, 0; green, 0; blue, 0 }  ,draw opacity=1 ]   (77,52.5) -- (84,52.5) ;

\draw [color={rgb, 255:red, 0; green, 0; blue, 0 }  ,draw opacity=1 ]   (91,52.5) -- (98,52.5) ;

\draw [color={rgb, 255:red, 0; green, 0; blue, 0 }  ,draw opacity=1 ]   (105,52.5) -- (112,52.5) ;

\draw [color={rgb, 255:red, 0; green, 0; blue, 0 }  ,draw opacity=1 ]   (119,52.5) -- (126,52.5) ;

\draw [color={rgb, 255:red, 0; green, 0; blue, 0 }  ,draw opacity=1 ]   (133,52.5) -- (140,52.5) ;

\draw [color={rgb, 255:red, 0; green, 0; blue, 0 }  ,draw opacity=1 ]   (147,52.5) -- (154,52.5) ;

\draw [color={rgb, 255:red, 0; green, 0; blue, 0 }  ,draw opacity=1 ]   (161,52.5) -- (168,52.5) ;

\draw [color={rgb, 255:red, 0; green, 0; blue, 0 }  ,draw opacity=1 ]   (210,52.5) -- (217,52.5) ;

\draw [color={rgb, 255:red, 0; green, 0; blue, 0 }  ,draw opacity=1 ]   (224,52.5) -- (231,52.5) ;

\draw [color={rgb, 255:red, 0; green, 0; blue, 0 }  ,draw opacity=1 ]   (238,52.5) -- (245,52.5) ;

\draw [color={rgb, 255:red, 0; green, 0; blue, 0 }  ,draw opacity=1 ] [dash pattern={on 0.84pt off 2.51pt}]  (168,66.5) -- (196,66.5) ;

\draw  [color={rgb, 255:red, 0; green, 0; blue, 0 }  ,draw opacity=1 ] (35,66.5) .. controls (35,64.57) and (36.57,63) .. (38.5,63) .. controls (40.43,63) and (42,64.57) .. (42,66.5) .. controls (42,68.43) and (40.43,70) .. (38.5,70) .. controls (36.57,70) and (35,68.43) .. (35,66.5) -- cycle ;
\draw  [color={rgb, 255:red, 0; green, 0; blue, 0 }  ,draw opacity=1 ] (49,66.5) .. controls (49,64.57) and (50.57,63) .. (52.5,63) .. controls (54.43,63) and (56,64.57) .. (56,66.5) .. controls (56,68.43) and (54.43,70) .. (52.5,70) .. controls (50.57,70) and (49,68.43) .. (49,66.5) -- cycle ;
\draw  [color={rgb, 255:red, 0; green, 0; blue, 0 }  ,draw opacity=1 ] (63,66.5) .. controls (63,64.57) and (64.57,63) .. (66.5,63) .. controls (68.43,63) and (70,64.57) .. (70,66.5) .. controls (70,68.43) and (68.43,70) .. (66.5,70) .. controls (64.57,70) and (63,68.43) .. (63,66.5) -- cycle ;
\draw  [color={rgb, 255:red, 0; green, 0; blue, 0 }  ,draw opacity=1 ] (77,66.5) .. controls (77,64.57) and (78.57,63) .. (80.5,63) .. controls (82.43,63) and (84,64.57) .. (84,66.5) .. controls (84,68.43) and (82.43,70) .. (80.5,70) .. controls (78.57,70) and (77,68.43) .. (77,66.5) -- cycle ;
\draw  [color={rgb, 255:red, 0; green, 0; blue, 0 }  ,draw opacity=1 ] (91,66.5) .. controls (91,64.57) and (92.57,63) .. (94.5,63) .. controls (96.43,63) and (98,64.57) .. (98,66.5) .. controls (98,68.43) and (96.43,70) .. (94.5,70) .. controls (92.57,70) and (91,68.43) .. (91,66.5) -- cycle ;
\draw  [color={rgb, 255:red, 0; green, 0; blue, 0 }  ,draw opacity=1 ] (105,66.5) .. controls (105,64.57) and (106.57,63) .. (108.5,63) .. controls (110.43,63) and (112,64.57) .. (112,66.5) .. controls (112,68.43) and (110.43,70) .. (108.5,70) .. controls (106.57,70) and (105,68.43) .. (105,66.5) -- cycle ;
\draw  [color={rgb, 255:red, 0; green, 0; blue, 0 }  ,draw opacity=1 ] (119,66.5) .. controls (119,64.57) and (120.57,63) .. (122.5,63) .. controls (124.43,63) and (126,64.57) .. (126,66.5) .. controls (126,68.43) and (124.43,70) .. (122.5,70) .. controls (120.57,70) and (119,68.43) .. (119,66.5) -- cycle ;
\draw  [color={rgb, 255:red, 0; green, 0; blue, 0 }  ,draw opacity=1 ] (133,66.5) .. controls (133,64.57) and (134.57,63) .. (136.5,63) .. controls (138.43,63) and (140,64.57) .. (140,66.5) .. controls (140,68.43) and (138.43,70) .. (136.5,70) .. controls (134.57,70) and (133,68.43) .. (133,66.5) -- cycle ;
\draw  [color={rgb, 255:red, 0; green, 0; blue, 0 }  ,draw opacity=1 ] (147,66.5) .. controls (147,64.57) and (148.57,63) .. (150.5,63) .. controls (152.43,63) and (154,64.57) .. (154,66.5) .. controls (154,68.43) and (152.43,70) .. (150.5,70) .. controls (148.57,70) and (147,68.43) .. (147,66.5) -- cycle ;
\draw  [color={rgb, 255:red, 0; green, 0; blue, 0 }  ,draw opacity=1 ] (161,66.5) .. controls (161,64.57) and (162.57,63) .. (164.5,63) .. controls (166.43,63) and (168,64.57) .. (168,66.5) .. controls (168,68.43) and (166.43,70) .. (164.5,70) .. controls (162.57,70) and (161,68.43) .. (161,66.5) -- cycle ;
\draw [color={rgb, 255:red, 0; green, 0; blue, 0 }  ,draw opacity=1 ]   (42,66.5) -- (49,66.5) ;

\draw [color={rgb, 255:red, 0; green, 0; blue, 0 }  ,draw opacity=1 ]   (56,66.5) -- (63,66.5) ;

\draw [color={rgb, 255:red, 0; green, 0; blue, 0 }  ,draw opacity=1 ]   (70,66.5) -- (77,66.5) ;

\draw [color={rgb, 255:red, 0; green, 0; blue, 0 }  ,draw opacity=1 ]   (84,66.5) -- (91,66.5) ;

\draw [color={rgb, 255:red, 0; green, 0; blue, 0 }  ,draw opacity=1 ]   (98,66.5) -- (105,66.5) ;

\draw [color={rgb, 255:red, 0; green, 0; blue, 0 }  ,draw opacity=1 ]   (112,66.5) -- (119,66.5) ;

\draw [color={rgb, 255:red, 0; green, 0; blue, 0 }  ,draw opacity=1 ]   (126,66.5) -- (133,66.5) ;

\draw [color={rgb, 255:red, 0; green, 0; blue, 0 }  ,draw opacity=1 ]   (140,66.5) -- (147,66.5) ;

\draw [color={rgb, 255:red, 0; green, 0; blue, 0 }  ,draw opacity=1 ]   (154,66.5) -- (161,66.5) ;

\draw  [color={rgb, 255:red, 0; green, 0; blue, 0 }  ,draw opacity=1 ] (210,66.5) .. controls (210,64.57) and (211.57,63) .. (213.5,63) .. controls (215.43,63) and (217,64.57) .. (217,66.5) .. controls (217,68.43) and (215.43,70) .. (213.5,70) .. controls (211.57,70) and (210,68.43) .. (210,66.5) -- cycle ;
\draw  [color={rgb, 255:red, 0; green, 0; blue, 0 }  ,draw opacity=1 ] (224,66.5) .. controls (224,64.57) and (225.57,63) .. (227.5,63) .. controls (229.43,63) and (231,64.57) .. (231,66.5) .. controls (231,68.43) and (229.43,70) .. (227.5,70) .. controls (225.57,70) and (224,68.43) .. (224,66.5) -- cycle ;
\draw  [color={rgb, 255:red, 0; green, 0; blue, 0 }  ,draw opacity=1 ] (238,66.5) .. controls (238,64.57) and (239.57,63) .. (241.5,63) .. controls (243.43,63) and (245,64.57) .. (245,66.5) .. controls (245,68.43) and (243.43,70) .. (241.5,70) .. controls (239.57,70) and (238,68.43) .. (238,66.5) -- cycle ;
\draw [color={rgb, 255:red, 0; green, 0; blue, 0 }  ,draw opacity=1 ]   (217,66.5) -- (224,66.5) ;

\draw [color={rgb, 255:red, 0; green, 0; blue, 0 }  ,draw opacity=1 ]   (231,66.5) -- (238,66.5) ;

\draw    (38.5,63) .. controls (42.5,58.67) and (63.17,60) .. (66.5,63) ;

\draw    (52.5,63) .. controls (56.5,58.67) and (77.17,60) .. (80.5,63) ;

\draw    (66.5,63) .. controls (70.5,58.67) and (91.17,60) .. (94.5,63) ;

\draw    (80.5,63) .. controls (84.5,58.67) and (105.17,60) .. (108.5,63) ;

\draw    (94.5,63) .. controls (98.5,58.67) and (119.17,60) .. (122.5,63) ;

\draw    (108.5,63) .. controls (112.5,58.67) and (133.17,60) .. (136.5,63) ;

\draw    (122.5,63) .. controls (126.5,58.67) and (147.17,60) .. (150.5,63) ;

\draw    (136.5,63) .. controls (140.5,58.67) and (161.17,60) .. (164.5,63) ;

\draw  [dash pattern={on 0.84pt off 2.51pt}]  (150.5,63) .. controls (154.5,58.67) and (175.17,60) .. (178.5,63) ;

\draw    (213.5,63) .. controls (217.5,58.67) and (238.17,60) .. (241.5,63) ;

\draw    (199.5,63) .. controls (203.5,58.67) and (224.17,60) .. (227.5,63) ;

\draw    (56,35) -- (30,35) ;
\draw [shift={(28,35)}, rotate = 360] [color={rgb, 255:red, 0; green, 0; blue, 0 }  ][line width=0.75]    (10.93,-3.29) .. controls (6.95,-1.4) and (3.31,-0.3) .. (0,0) .. controls (3.31,0.3) and (6.95,1.4) .. (10.93,3.29)   ;

\draw    (84,35) -- (250,35) ;
\draw [shift={(252,35)}, rotate = 180] [color={rgb, 255:red, 0; green, 0; blue, 0 }  ][line width=0.75]    (10.93,-3.29) .. controls (6.95,-1.4) and (3.31,-0.3) .. (0,0) .. controls (3.31,0.3) and (6.95,1.4) .. (10.93,3.29)   ;

\draw  [color={rgb, 255:red, 0; green, 0; blue, 0 }  ,draw opacity=1 ] (133,80.5) .. controls (133,78.57) and (134.57,77) .. (136.5,77) .. controls (138.43,77) and (140,78.57) .. (140,80.5) .. controls (140,82.43) and (138.43,84) .. (136.5,84) .. controls (134.57,84) and (133,82.43) .. (133,80.5) -- cycle ;
\draw  [color={rgb, 255:red, 0; green, 0; blue, 0 }  ,draw opacity=1 ] (147,80.5) .. controls (147,78.57) and (148.57,77) .. (150.5,77) .. controls (152.43,77) and (154,78.57) .. (154,80.5) .. controls (154,82.43) and (152.43,84) .. (150.5,84) .. controls (148.57,84) and (147,82.43) .. (147,80.5) -- cycle ;
\draw  [color={rgb, 255:red, 0; green, 0; blue, 0 }  ,draw opacity=1 ] (161,80.5) .. controls (161,78.57) and (162.57,77) .. (164.5,77) .. controls (166.43,77) and (168,78.57) .. (168,80.5) .. controls (168,82.43) and (166.43,84) .. (164.5,84) .. controls (162.57,84) and (161,82.43) .. (161,80.5) -- cycle ;
\draw  [color={rgb, 255:red, 0; green, 0; blue, 0 }  ,draw opacity=1 ] (210,80.5) .. controls (210,78.57) and (211.57,77) .. (213.5,77) .. controls (215.43,77) and (217,78.57) .. (217,80.5) .. controls (217,82.43) and (215.43,84) .. (213.5,84) .. controls (211.57,84) and (210,82.43) .. (210,80.5) -- cycle ;
\draw  [color={rgb, 255:red, 0; green, 0; blue, 0 }  ,draw opacity=1 ] (224,80.5) .. controls (224,78.57) and (225.57,77) .. (227.5,77) .. controls (229.43,77) and (231,78.57) .. (231,80.5) .. controls (231,82.43) and (229.43,84) .. (227.5,84) .. controls (225.57,84) and (224,82.43) .. (224,80.5) -- cycle ;
\draw [color={rgb, 255:red, 0; green, 0; blue, 0 }  ,draw opacity=1 ] [dash pattern={on 0.84pt off 2.51pt}]  (168,80.5) -- (196,80.5) ;

\draw    (52.5,77) .. controls (58.75,72) and (99.25,71) .. (108.5,77) ;

\draw  [color={rgb, 255:red, 0; green, 0; blue, 0 }  ,draw opacity=1 ] (91,94.5) .. controls (91,92.57) and (92.57,91) .. (94.5,91) .. controls (96.43,91) and (98,92.57) .. (98,94.5) .. controls (98,96.43) and (96.43,98) .. (94.5,98) .. controls (92.57,98) and (91,96.43) .. (91,94.5) -- cycle ;
\draw  [color={rgb, 255:red, 0; green, 0; blue, 0 }  ,draw opacity=1 ] (105,94.5) .. controls (105,92.57) and (106.57,91) .. (108.5,91) .. controls (110.43,91) and (112,92.57) .. (112,94.5) .. controls (112,96.43) and (110.43,98) .. (108.5,98) .. controls (106.57,98) and (105,96.43) .. (105,94.5) -- cycle ;
\draw  [color={rgb, 255:red, 0; green, 0; blue, 0 }  ,draw opacity=1 ] (119,94.5) .. controls (119,92.57) and (120.57,91) .. (122.5,91) .. controls (124.43,91) and (126,92.57) .. (126,94.5) .. controls (126,96.43) and (124.43,98) .. (122.5,98) .. controls (120.57,98) and (119,96.43) .. (119,94.5) -- cycle ;
\draw  [color={rgb, 255:red, 0; green, 0; blue, 0 }  ,draw opacity=1 ] (133,94.5) .. controls (133,92.57) and (134.57,91) .. (136.5,91) .. controls (138.43,91) and (140,92.57) .. (140,94.5) .. controls (140,96.43) and (138.43,98) .. (136.5,98) .. controls (134.57,98) and (133,96.43) .. (133,94.5) -- cycle ;
\draw  [color={rgb, 255:red, 0; green, 0; blue, 0 }  ,draw opacity=1 ] (147,94.5) .. controls (147,92.57) and (148.57,91) .. (150.5,91) .. controls (152.43,91) and (154,92.57) .. (154,94.5) .. controls (154,96.43) and (152.43,98) .. (150.5,98) .. controls (148.57,98) and (147,96.43) .. (147,94.5) -- cycle ;
\draw  [color={rgb, 255:red, 0; green, 0; blue, 0 }  ,draw opacity=1 ] (161,94.5) .. controls (161,92.57) and (162.57,91) .. (164.5,91) .. controls (166.43,91) and (168,92.57) .. (168,94.5) .. controls (168,96.43) and (166.43,98) .. (164.5,98) .. controls (162.57,98) and (161,96.43) .. (161,94.5) -- cycle ;
\draw [color={rgb, 255:red, 40; green, 127; blue, 229 }  ,draw opacity=1 ]   (52.5,84) -- (80.5,91) ;

\draw [color={rgb, 255:red, 40; green, 127; blue, 229 }  ,draw opacity=1 ]   (66.5,84) -- (80.5,91) ;

\draw [color={rgb, 255:red, 40; green, 127; blue, 229 }  ,draw opacity=1 ]   (80.5,84) -- (80.5,91) ;

\draw [color={rgb, 255:red, 40; green, 127; blue, 229 }  ,draw opacity=1 ]   (94.5,84) -- (80.5,91) ;

\draw [color={rgb, 255:red, 40; green, 127; blue, 229 }  ,draw opacity=1 ]   (108.5,84) -- (80.5,91) ;

\draw  [color={rgb, 255:red, 0; green, 0; blue, 0 }  ,draw opacity=1 ] (77,94.5) .. controls (77,92.57) and (78.57,91) .. (80.5,91) .. controls (82.43,91) and (84,92.57) .. (84,94.5) .. controls (84,96.43) and (82.43,98) .. (80.5,98) .. controls (78.57,98) and (77,96.43) .. (77,94.5) -- cycle ;
\draw  [dash pattern={on 0.84pt off 2.51pt}]  (80.5,91) .. controls (87.25,85) and (172.75,85) .. (182,91) ;

\draw  [color={rgb, 255:red, 0; green, 0; blue, 0 }  ,draw opacity=1 ] (49,80.5) .. controls (49,78.57) and (50.57,77) .. (52.5,77) .. controls (54.43,77) and (56,78.57) .. (56,80.5) .. controls (56,82.43) and (54.43,84) .. (52.5,84) .. controls (50.57,84) and (49,82.43) .. (49,80.5) -- cycle ;
\draw  [color={rgb, 255:red, 0; green, 0; blue, 0 }  ,draw opacity=1 ] (63,80.5) .. controls (63,78.57) and (64.57,77) .. (66.5,77) .. controls (68.43,77) and (70,78.57) .. (70,80.5) .. controls (70,82.43) and (68.43,84) .. (66.5,84) .. controls (64.57,84) and (63,82.43) .. (63,80.5) -- cycle ;
\draw  [color={rgb, 255:red, 0; green, 0; blue, 0 }  ,draw opacity=1 ] (77,80.5) .. controls (77,78.57) and (78.57,77) .. (80.5,77) .. controls (82.43,77) and (84,78.57) .. (84,80.5) .. controls (84,82.43) and (82.43,84) .. (80.5,84) .. controls (78.57,84) and (77,82.43) .. (77,80.5) -- cycle ;
\draw  [color={rgb, 255:red, 0; green, 0; blue, 0 }  ,draw opacity=1 ] (91,80.5) .. controls (91,78.57) and (92.57,77) .. (94.5,77) .. controls (96.43,77) and (98,78.57) .. (98,80.5) .. controls (98,82.43) and (96.43,84) .. (94.5,84) .. controls (92.57,84) and (91,82.43) .. (91,80.5) -- cycle ;
\draw  [color={rgb, 255:red, 0; green, 0; blue, 0 }  ,draw opacity=1 ] (105,80.5) .. controls (105,78.57) and (106.57,77) .. (108.5,77) .. controls (110.43,77) and (112,78.57) .. (112,80.5) .. controls (112,82.43) and (110.43,84) .. (108.5,84) .. controls (106.57,84) and (105,82.43) .. (105,80.5) -- cycle ;
\draw  [color={rgb, 255:red, 0; green, 0; blue, 0 }  ,draw opacity=1 ] (119,80.5) .. controls (119,78.57) and (120.57,77) .. (122.5,77) .. controls (124.43,77) and (126,78.57) .. (126,80.5) .. controls (126,82.43) and (124.43,84) .. (122.5,84) .. controls (120.57,84) and (119,82.43) .. (119,80.5) -- cycle ;
\draw  [color={rgb, 255:red, 0; green, 0; blue, 0 }  ,draw opacity=1 ] (196,94.5) .. controls (196,92.57) and (197.57,91) .. (199.5,91) .. controls (201.43,91) and (203,92.57) .. (203,94.5) .. controls (203,96.43) and (201.43,98) .. (199.5,98) .. controls (197.57,98) and (196,96.43) .. (196,94.5) -- cycle ;
\draw  [color={rgb, 255:red, 0; green, 0; blue, 0 }  ,draw opacity=1 ] (196,80.5) .. controls (196,78.57) and (197.57,77) .. (199.5,77) .. controls (201.43,77) and (203,78.57) .. (203,80.5) .. controls (203,82.43) and (201.43,84) .. (199.5,84) .. controls (197.57,84) and (196,82.43) .. (196,80.5) -- cycle ;
\draw  [color={rgb, 255:red, 0; green, 0; blue, 0 }  ,draw opacity=1 ] (196,66.5) .. controls (196,64.57) and (197.57,63) .. (199.5,63) .. controls (201.43,63) and (203,64.57) .. (203,66.5) .. controls (203,68.43) and (201.43,70) .. (199.5,70) .. controls (197.57,70) and (196,68.43) .. (196,66.5) -- cycle ;
\draw     ;

\draw [color={rgb, 255:red, 40; green, 127; blue, 229 }  ,draw opacity=1 ]   (80.5,98) -- (137,105) ;

\draw  [color={rgb, 255:red, 0; green, 0; blue, 0 }  ,draw opacity=1 ] (133,108.5) .. controls (133,106.57) and (134.57,105) .. (136.5,105) .. controls (138.43,105) and (140,106.57) .. (140,108.5) .. controls (140,110.43) and (138.43,112) .. (136.5,112) .. controls (134.57,112) and (133,110.43) .. (133,108.5) -- cycle ;

\draw (69.5,31.5) node   {$L$};
\draw (21.5,65.4) node [scale=0.8]  {$-1$};
\draw (21.5,80.2) node [scale=0.8]  {$-3$};
\draw (21.5,94.2) node [scale=0.8]  {$-7$};
\draw (23,107.8) node [scale=0.8]  {$-15$};

\end{tikzpicture}
\caption{In this diagram we consider the pooling operation applied to a 1D chain. The 2D case, that of images, only differs in that the process occurs in both directions in parallel -- the pools remain the same size. Nodes and edges are shown in black, blue indicates how the nodes are pooled into the cliques with the next chain being the result of the pooling operation. In this way we can illustrate four iterations of the pooling operation. In the first the cliques are of size two and the chain reduces in length by one -- half from each end. In the second the cliques are of size three, due to the greater reach of the inherited edge connections, and the chain has reduced by two more, one from each end, for a total of three. This process continues with the reach becoming greater, the size of the pools increasing and the chain reducing more rapidly, as indicated on the left.}
\label{1Dchaindiagram}
\end{center}
\vskip -0.2in
\end{figure}
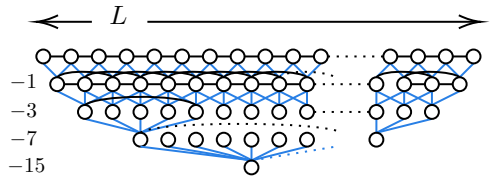
\raggedbottom

\begin{proof}
Figure \ref{1Dchaindiagram} illustrates how the pools grow over successive iterations. The trend appears to be $2^n-1$ and we can show that this is indeed the case. Consider the size of the cliques at a particular iteration, $c_i$, and the distance along the chain each node is connected, $d_i$. By inspection, each node is connected to every neighbour up to $d_i$ and so this group forms a clique of size
\begin{align*}
    c_i = d_i + 1
\end{align*}
with the additional 1 accounting for the node itself. Tracing the inheritance of connections into the next layer gives the connected distance of the pooled nodes as
\begin{align*}
    d_{i+1} &= \tfrac{1}{2}(c_i - 1) + d_i + \tfrac{1}{2}(c_i - 1)\\
            &= c_i - 1 + d_i = 2d_i.
\end{align*}
Where we first go up one side of the clique, along the connected distance and then back down the side of another clique. So the distance does double each time with the cliques, and pools, being one more. The pooling routine as applied to images then, is to use pools of increasing size, with no padding, and a unit stride. Specifically, given the initial degree of 1 implied by the convolutions, $(2\times2),(3\times3),(5\times5),(9\times9)$\ldots$(2^{n-1}+1\times2^{n-1}+1)$

In addition to the simplicity of the operation, if we observe the reduction in the length a second benefit is made apparent. On the left we reduce by the left half of the pool, on the right by the right half, but not by the middle on either. So the length is reduced by $c_i-1 = d_i$. As $d_i$ doubles each time, starting from 1, the reduction after $n-$pools is
\begin{align*}
    r_i = \sum_{i = 1}^n 2^{n-1} = 2^n - 1.
\end{align*}

\end{proof}

\end{document}